\documentclass[doublesided,msc,ai,logo]{infthesis}
\pdfoutput=1

\usepackage[pageref]{backref}
\usepackage[colorlinks]{hyperref}
\usepackage{hyperref}
\usepackage{stmaryrd}
\usepackage{attrib}
\usepackage{amsmath}
\usepackage{covington}
\usepackage{nomencl}
\usepackage{algorithm}
\usepackage{algpseudocode}
\usepackage{tikz}
\usepackage{tikz-qtree}
\usepackage{tikz-qtree-compat}
\usepackage{tikz-dependency}
\usepackage{amssymb}
\usepackage{listings}
\usepackage{natbib}
\usepackage{multicol}
\usepackage{float}
\usetikzlibrary{shapes,arrows}

\tikzstyle{decision} = [diamond, draw, fill=blue!20, 
text width=4em, text badly centered, node distance=3cm, inner sep=0pt]
\tikzstyle{block} = [rectangle, draw, fill=blue!20, 
text width=5em, text centered, rounded corners, minimum height=3em]
\tikzstyle{line} = [draw, -latex']
\tikzstyle{cloud} = [draw, ellipse,fill=red!20, node distance=3cm,
minimum height=2em]

\hypersetup{
	colorlinks,
	citecolor=black,
	filecolor=black,
	linkcolor=black,
	urlcolor=black
}

\bibpunct{(}{)}{;}{a}{,}{,}

\makenomenclature

\title{CCG Parsing and \\Multiword Expressions}
\author{Miryam de Lhoneux}
\submityear{2014}

\pdfoutput=1

\newcommand \insertbibliography{\bibliography{bibliography}}

\newcommand{\tab}{\-\hspace{1cm}}

\newcommand{\citepos}[1]{\citeauthor{#1}'s (\citeyear{#1})}
\newcommand{\subscript}[1]{\raisebox{-.4ex}{\scriptsize #1}}
\newcommand{\modelB}{model\subscript{B} }
\newcommand{\modelA}{model\subscript{A} }

\renewcommand*{\backref}[1]{}
\renewcommand*{\backrefalt}[4]{%
	\ifcase #1 (Not cited.)%
	\or        (on p.~#2)%
	\else      (on pp.~#2)%
\fi}

\abstract{
	\tab This thesis presents a study about the integration of information about Multiword Expressions (MWEs) into parsing with Combinatory Categorial Grammar (CCG). We build on previous work by \citet{nivre2004} and by \citet{korkontzelos2010} who have shown the benefit of adding information about MWEs to syntactic parsing by implementing a similar pipeline with CCG parsing. More specifically, we collapse MWEs to one token in training and test data in CCGbank, a corpus automatically created by \citet{Hockenmaier2003} which contains sentences annotated  with CCG derivations. Our collapsing algorithm however can only deal with MWEs when they form a constituent in the data which is one of the limitations of our approach.\\
	\tab We study the effect of collapsing training and test data. A parsing effect can be obtained if collapsed data help the parser in its decisions and a training effect can be obtained if training on the collapsed data improves results. We also collapse the gold standard and show that our model significantly outperforms the baseline model on our gold standard, which indicates that there is a training effect. We show that the baseline model performs significantly better on our gold standard when the data are collapsed before parsing than when the data are collapsed after parsing which indicates that there is a parsing effect. We show that these results can lead to improved performance on the non-collapsed standard benchmark although we fail to show that it does so significantly. We conclude that despite the limited settings, there are noticeable improvements from using MWEs in parsing. We discuss ways in which the incorporation of MWEs into parsing can be improved and hypothesize that this will lead to more substantial results.\\
	\tab We finally show that turning the MWE recognition part of the pipeline into an experimental part is a useful thing to do as we obtain different results with different recognizers.
}

\begin{document}
\begin{preliminary}
	\maketitle
	\begin{acknowledgements}
	\tab First and foremost, I would like to thank my supervisors Mark Steedman and Omri Abend. I went to see Mark in January with the story of where I was coming from and where I wanted to go and I knew right away I had knocked on the right door. He attentively listened to my story, got me in touch with Omri and together they helped me connect the dots so my thesis nicely fits into the picture. Omri was a very enthusiastic supervisor full of interesting ideas and I admire the patience he had with me. I am grateful to both of them for all the discussions and for taking the time to proofread my drafts as well as insightfully commenting on them.\\
	\tab I would also like to thank Bharat Ram Ambati who was very helpful in giving me a hand on some technical issues.\\
	\tab I am not sure I would have made it through this dissertation let alone this master if it were not for my AT pals who I also wish to thank. This year of living together in Appleton tower has been a tough one but it has created wonderful friendships.\\
	\tab My other friends and family should also be thanked for having supported me through my master thesis madness for two years in a row, I promise this is the last one.
\end{acknowledgements}

	\standarddeclaration
	\tableofcontents

	\listoffigures

	\listoftables

	\printnomenclature

\end{preliminary}

\chapter{Introduction}
\label{introduction}

\section{Motivation}
\tab Syntactic parsing is the task of analysing the relationships between words in a sentence and thereby giving a compositional account of sentences. Syntactic parsing is considered a central task in Natural Language Processing (henceforth, NLP\nomenclature{NLP}{Natural Language Processing}) and is used in many applications. It is a central but difficult task partly because it requires a lot of annotated data. Ground-breaking research in the 1990s have shown impressive results on this task but the major efforts have been directed to parsing on a specific domain (news) and a specific language (Enlglish) and the problem is far from being solved.\\
\tab Multiword Expressions (henceforth MWE(s)\nomenclature{MWE(s)}{Multiword Expression(s)}) are increasingly receiving attention in NLP. They represent a wide variety of phenomena with different properties but are generally agreed to be a group of multiple lexemes which have some level of idiomaticity or irregularity \citep{Sagetal01}. They represent varied phenomena but they are all generally considered a problem for NLP tasks and they are often a problem for syntactic parsing. \\
\tab Recent research is showing that information about MWEs help the syntactic parsing task \citep{nivre2004,korkontzelos2010} and inversely, information about syntactic analysis helps MWE identification \citep{Greenetal13,Weller&Heid2010,martens&vandeghinste2010}. Working on either of the task by using information from the other has thus been proven to be a useful thing to do in improving either of the other task and adding MWE information to the syntactic parsing task has proven useful in that it has helped increasing parsing accuracy. Work on adding MWE information to syntactic parsing so far has been restricted to certain types of MWEs and hence leaves a lot of room for improvement.\\
\tab Combinatory Categorial Grammar (henceforth CCG\nomenclature{CCG}{Combinatory Categorial Grammar}) is a strongly lexicalized formalism that is increasingly being used for parsing in NLP applications because of its computational and linguistic properties and because it performs almost state-of-the-art on the relatively accurate parses we have available for it.\\ 
\tab For these reasons, CCG parsing is an ideal framework to carry on the work on the interaction between syntactic parsing and Multiword Expressions and because CCG is a lexicalized formalism and thus encodes a lot of information in the lexicon, it would be useful to work on it by providing it with information about MWEs. 

\section{Aims}
\tab No work so far has tried to use MWE information to improve CCG parsing which is what we intend to do in this thesis. Different approaches to using MWE information for improving syntactic parsing have been conducted so far with different syntactic models and we will be conducting one of them which will argued to be far from ideal but a necessary first step useful to build a sound baseline. The approach we will be conducting consists in altering training and test data, i.e. collapsing MWEs to one lexical item in them. By collapsing we mean grouping MWEs together so that they form one token, and hence retokenize the sentence. We will be using this term in the remainder of the thesis. A change in the approach we will propose is to make MWE recognition an experimental part of the pipeline so that we can find out whether or not conducting this type of approach on different types of MWEs lead to different results. \\
\tab The two research questions we will therefore try to answer are first whether or not we can improve CCG parsing with MWEs and second whether or not applying the same collapsing approach to different types of MWEs lead to different results.

\section{Overview}
\tab We will give an overview of the background literature to further support our motivations and elaborate on the research questions in Chapter \ref{relatedwork}. We will then explain and motivate the methodology we propose to use in order to answer the research questions in Chapter \ref{methodology}. We will present our experiments and results in Chapter \ref{exp} and discuss the results and limitations of the present approach in Chapter \ref{discussion}. We will conclude from our study and propose avenues of research in Chapter \ref{ccl}.

\chapter{Related Work}
\label{relatedwork}
\section{Introduction}
\label{intro}
\tab This chapter presents an overview of the research fields on which this thesis is based: Syntactic Parsing (Section \ref{syntacticparsing}), Multiword Expressions (Section \ref{MWEs}) and Combinatory Categorial Grammar (Section \ref{CCG}). For reasons of time and space, they are only briefly introduced with the background knowledge necessary for understanding the remainder of the thesis. Work on the interaction between the three frameworks is then presented in Sections \ref{syntacticparsingandMWES} (between Multiword Expressions and Syntactic Parsing) and \ref{CCGparsing} (between CCG and Syntactic Parsing). The objectives and research questions of the thesis are presented in Section \ref{relworkccl}.
\section{Syntactic Parsing}
\label{syntacticparsing}
\tab As said in the introduction, syntactic parsing is the task of analysing the relationships between words in a sentence and thereby giving a compositional account of sentences. It is considered by many (e.g. \citet{Clark2010}) to be a central task in NLP, it is used in many applications such as Tree-based Machine Translation, Question Answering etc. and is a first step towards semantic analysis. This compositional account can take the form of a Phrase Structure Tree or a Dependency Graph.\\
\-\hspace{1cm} A syntactic parser has two essential components: a grammar which determines the set of possible sentences in a language and an algorithm that determines the grammar rules that are applied for each word in the sentence. Early research \citep{ChurchandPatil82} has shown that existing grammars are highly ambiguous and has called for the need for two more components: a probability model for scoring the different possible parses and an algorithm for choosing the best scoring parse. Most parsing models include these 4 components.  \\
\tab Most work on syntactic parsing is based on a manually annotated corpus, the Penn Treebank \citep{PTB}\nomenclature{PTB}{Penn Treebank} which has been used to infer a probabilistic model together with a Context Free Grammar (henceforth CFG\nomenclature{CFG}{Context Free Grammar}) or a PCFG \nomenclature{PCFG}{Probabilistic Context Free Grammar}. There is ongoing work which is an exception to that tendency: work on parsing with precision grammars, i.e. hand-crafted grammars which are written so as to distinguish grammatical from ungrammatical sentences, such as Head-Driven Phrase Structure Grammar (henceforth HPSG\nomenclature{HPSG}{Head-Driven Phrase Structure Grammar}). The CKY algorithm, a chart-based algorithm has been widely used to determine the possible parse trees by building them bottom-up, i.e. from the leaf nodes to the sentence node. Ground-breaking research on the PTB was made in the late 90s for example by Collins \citep{Collins96, collins1997, Collins2003}.  In his seminal work, Collins formally defined the parsing problem as in Equation~\ref{eq:parsing}, in which Tbest is the best scoring parse tree among all the parse trees T for the sentence S. This is called a generative model: it models the sentence and the tree in a joint way and heavily relies on independence assumptions: it decomposes the probability of the tree into probabilities of rule application.
\begin{equation} \label{eq:parsing}
	Tbest = \underset{T}{\operatorname{argmax}} P(T,S)
\end{equation}
\tab Syntactic parsing has traditionally been evaluated by computing parsing accuracy and parsing accuracy has traditionally been measured with the PARSEVAL metric \citep{Parseval}. This parsing accuracy measure is obtained by comparing output parse trees with the treebank gold standard: labelled precision and recall of nodes in the trees are computed as well as their F-measure. Collins reported parsing accuracy results of over 90\% which were then slightly improved by \citet{charniak&johnson2005}. Since these early results however, a lot has been tried but not a lot of improvement has been made which highlights the difficulty of the task at hand. However, as pointed out by \citet{Clark2010}, research has shifted the focus away from building complex generative models to discriminative models, which estimate the probability of a parse tree for a sentence directly instead of estimating parse tree and sentence together, and which avoid independence assumptions. Research has also drawn attention to building grammars based on expressive formalisms such as Tree Adjoining Grammar (henceforth TAG\nomenclature{TAG}{Tree Adjoining Grammar}), Lexical Functional Grammar (henceforth LFG\nomenclature{LFG}{Lexical Functional Grammar}), HPSG and CCG which offer more linguistically accurate representations of language. According to \citet{nivre2010}, the tendency of using supervised methods for parsing is moving towards unsupervised methods such as in \citet{Spitkovsky}. Some authors have given up on improving parsing accuracy directly but have tried working on it indirectly by working on it as a joint task together with another task such such as \citet{Burkett2008} who show the benefit of combining Machine Translation and Syntactic Parsing as well as the work which has tried to improve Multiword Expression identification together with syntactic parsing, described in Section \ref{syntacticparsingandMWES}. \\
\-\hspace{1cm} 	In any case, although the percentages obtained by the seminal works on generative models may seem very high, syntactic parsing is far from being a solved problem, as argued by \citet{Clark2010}, because these percentages are inflated by recurring derivations and cannot deal with many important constructions. In addition, most work is focused on one domain and one language and need adaptation to other domains and languages which the lack of data make a problem hard to solve. Tremendous work is therefore still needed for syntactic parsing and the different shifts in research focus mentioned may still have a lot to offer.

\section{Multiword Expressions}
\label{MWEs}
\-\hspace{1cm} Multiword Expressions is an umbrella term that has been used to characterize a wide variety of phenomena. The most commonly acknowledged definition of this term since \citet{Sagetal01} is that it is a group of multiple lexemes which have some level of idiomaticity or irregularity. The multiple lexemes in a MWE will be called MWE units in the remainder of this thesis for convenience. This idiomaticity may be lexico-syntactic such as in the unusual coordination of a preposition and an adjective in `by and large'. It may be semantic such as in the idiom `kick the bucket' in which the meaning of the whole is not dividable into the meaning of the parts. It may be pragmatic such as in "good morning" which has a meaning attached to the situation in which it is said. Finally, it may be statistical such as the collocation `strong coffee' in which both units occur more frequently than expected.\\
\-\hspace{1cm} MWEs are generally also agreed to display different properties. They vary in flexibility: words may appear between the units of a flexible collocation (strong home-made coffee for example) but not between the units of a lexically fixed figurative expression such as `it's raining cats and dogs'. They also vary in compositionality: `Strong coffee' is fully compositional whereas `kick the bucket' is not and `spill the beans' is semantically decomposable, i.e. the meaning of the whole is not predictable from the meaning of the parts but can be decomposed into its parts: if `spill' is interpreted as `reveal' and `the beans' as `the secret' \citep{nunberg1994}.
Despite these varied properties they are all generally agreed to be a pain in the neck for NLP applications and the importance of dealing with them properly has been increasing over the past decade. As described at length in \citepos{Kim2008} thesis, `dealing with' MWEs consists in developing systems and models for various kinds of tasks. For syntactic analysis, it is important to identify them in text and extract them to a dictionary. For semantic understanding, it is important to measure their compositionality, classify them and interpret them.

\section{Combinatory Categorial Grammar}
\label{CCG}
\begin{figure}[H]
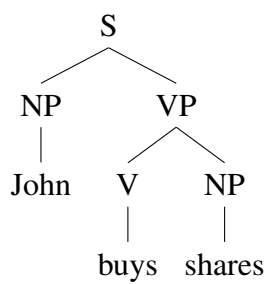

		\Tree [.S [.NP John ] [.VP [.V buys ] [.NP shares ] ] ]
		\caption{Traditional tree\label{fig:1}}
	\end{figure}

\-\hspace{1cm}Combinatory Categorial Grammar \citep{Steedman2000} is a strongly lexicalized grammar formalism which is currently gaining popularity in the NLP community. It was built with the intent of being linguistically aware as well as computationally tractable partly as a reaction to transformationalist ideas which were predominant in formal grammars at the time. It differs mainly from these in having one component including syntactic and semantic information instead of having separate modules for each in the grammar. Similarly, instead of having a large amount of rules and a lexicon as is the case in traditional grammars, it has a small set of universal rules and a lexicon which encodes most syntactic information.
For the sentence `John buys shares', a traditional grammar has information in the lexicon: that John is an NP, that buys is a verb, that shares is an NP, and in the grammar: that a V and an NP form a VP and that an NP and a VP form a sentence S, as in Figure \ref{fig:1}.
By contrast, for the same sentence, CCG has information in its lexicon that John is an NP, that shares is an NP and that buys first takes an NP to its right then an NP to its left to form a sentence S, as in Figure \ref{fig:2}.
\begin{multicols}{2}
	\columnbreak
	\begin{figure}[H]
		\Tree [.S [.NP John ] [.S\textbackslash NP [.(S\textbackslash NP)/NP buys ] [.NP shares ] ] ]
		\caption{CCG tree\label{fig:2}}
	\end{figure}
\end{multicols}

\begin{figure}[H]
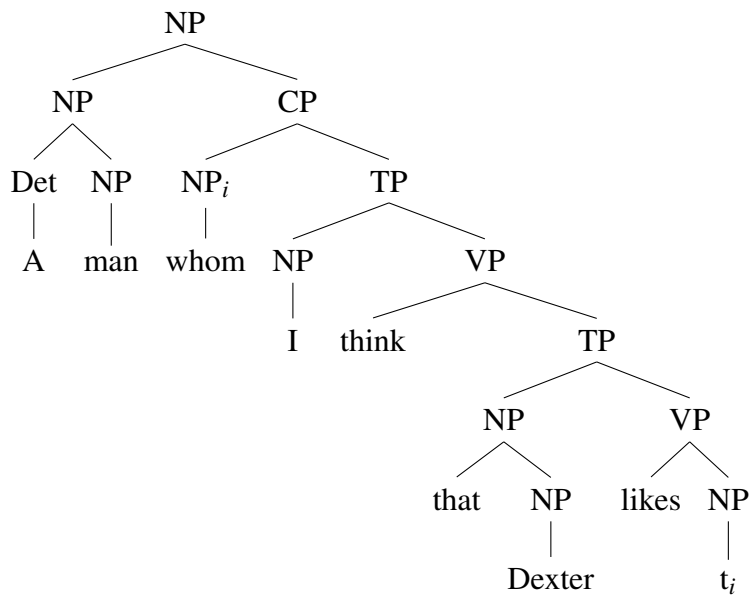

	\Tree [.NP [.NP [.Det A ] [.NP man ] ] [.CP
			[.NP_i whom ] [.TP
				[.NP  I  ] [.VP think  [.TP
	[.NP that [.NP Dexter ] ] [.VP likes [.NP t_i ] ] ] ] ] ] ] 
	\caption{Traditional tree\label{fig:3}}
\end{figure}

\tab I will not go into details on how this exactly works but lexical categories work either as functor or as argument and a set of combinatory rules allow them to combine. For example, the category (S\textbackslash NP)/NP works as a functor that takes an NP to the right (indicated by the forward slash followed by an NP). The category of `buys' therefore can combine with the category of `shares' to result in the category S\textbackslash NP which in turns takes an NP argument to the left (indicated by the backslash followed by an NP), which it finds in the category of `John' to form a sentence S.\\
\tab This grammar architecture allows CCG to deal elegantly with long-range dependencies. Instead of adding a level of representation in the form of a trace as in Figure \ref{fig:3}, the grammar has universal rules which allow the combination of lexical items, as shown in Figure \ref{fig:4}. This has computational advantages and linguistic plausibility: Linguistics is increasingly adopting a view of grammar where syntax and the lexicon are intertwined. It is a tenet of the recently emerging framework of Construction Grammar \citep{Hoffmann2013}. These linguistic and computational properties have made it a widely used framework across NLP research.

\begin{figure}[H]
	\Tree [.NP [.NP [.NP/N A ] [.N man ] ]  [.NP\textbackslash NP
			[.(NP\textbackslash NP)/(S/NP) whom ] [.S/NP
				[.S/(S\textbackslash NP)  [.NP I ]  ] [.(S\textbackslash NP)/NP [.(S\textbackslash NP)/S think ]  [.S/NP
	[.S/(S\textbackslash NP) [.NP [.NP/N that ] [.N Dexter ] ] ] [.(S\textbackslash NP)/NP likes ] ] ] ] ] ]
	\caption{CCG tree\label{fig:4}} 
\end{figure}
\section{Syntactic Parsing and Multiword Expressions}
\label{syntacticparsingandMWES}

\-\hspace{1cm}As mentioned in Section \ref{MWEs}, the identification of MWEs is important for syntactic analysis. Because they have unusual properties however, their analysis can be quite problematic. The question of how to deal with MWEs for syntactic parsing has been raised by many. It has been approached in different ways. Researchers working with precision grammars such as HPSG for example have accommodated the lexical entries for MWEs in the lexicon so that MWEs are not a problem for parsing. Researchers on data-induced grammars have accommodated the testing and/or training data before parsing. Recent research has proposed to both change the lexicon and the parsing algorithm. I will briefly describe each of these approaches in turn. Because I will be using the second approach for reasons explained in \ref{choiceofapp}, I will describe it in a more detailed manner than the other two. \\

\subsection{Changing the lexicon}
\tab Different types of lexical entries have been proposed for MWEs in the grammar. A lot of research proposes to simply analyse all MWEs as `words-with-spaces', i.e. group the MWE units together in the syntactic analysis. This analysis has been argued against by many. \citet{Sagetal01} have suggested sophisticated ways of representing the different MWE types in a grammar, which have been partly implemented within the framework of the precision grammar HPSG, as described by \citet{Copestakeetal2002}. \citet{Zhangetal2006} recently established that MWEs are a tremendous source of parse failures when parsing with a precision grammar such as HPSG and henceforth proposed a way of using this information to identify new MWEs and enrich a lexicon: they suggested using parse failures to predict the existence of a MWE.

\subsection{Changing the data}
\label{changingdata}
\tab Since the seminal work of \citet{nivre2004}, research has shown that treating MWEs as one token or a `word-with-spaces' in test and/or in 	training data before parsing and/or training leads to an improvement in parsing accuracy. \citet{nivre2004} have shown that to be true for deterministic dependency parsing and \citet{korkontzelos2010} have shown that to be true for shallow parsing. 
\tab The two approaches are quite different and I will describe each in turn.
\subsubsection{Changing training and test data}
\tab \citet{nivre2004} created two versions of a treebank, one in which MWEs are annotated as if compositional and one in which they are joined as one lexical item. They have shown that training a parser on the second version of the treebank leads to a better parsing accuracy. They used a corpus with manual MWE annotation to create both versions of the treebank and hence simulated `perfect' MWE recognition. MWE annotation however only consists in a few MWE types so it is not comprehensive. They reported improvement in parsing accuracy of the MWEs themselves but also of their surrounding syntactic structure. They opened the gate for improving syntactic parsing with MWE information but left many questions unanswered: whether or not their results port to other syntactic parsing models, whether or not the full potential they obtained with `perfect' recognition of MWEs can be obtained with an automatic recognizer and whether or not this potential can be increased when recognizing other types of MWEs. \citet{Constant2012} have recently shown that with an automatic recognizer, the parsing accuracy improvement is not as dramatic as predicted by \citet{nivre2004}, giving a negative answer to the second of these questions.
\subsubsection{Changing test data}
\label{korkontz}
\tab \citet{korkontzelos2010} reported similar parsing accuracy improvements for shallow parsing, showing that \citet{nivre2004}'s results do seem to port to at least one other parsing model. Their technique is however quite different. They created a corpus containing a large number of pre-selected MWEs (randomly chosen from WordNet) and converted it to a version in which the MWE units are collapsed to one lexical item. They POS-tagged the two versions of the corpus before parsing each. They consequently analyzed the differences in output. In order to do so, they randomly selected a sample of output from both parsed corpora and built a taxonomy of changes they observed from one to the other. For each class in the taxonomy, they determined whether the change in output led to increased accuracy, decreased accuracy or did not change the accuracy. They automatically classified the rest of the output data and observed an overall increase in accuracy. Their work not only confirmed the results obtained from previous work but also provided an insightful qualitative analysis of changes obtained with their method. They believe the improvement in accuracy is partly due to the fact that the parsing model backs off to POS-tags for rare and unseen words. Collapsed MWEs are not known by the parser but they are still mostly assigned a sensible POS-tag because the POS-tagger uses contextual information. 

\subsection{Changing the lexicon and the parsing algorithm}
\tab A lot of work has shown that although MWE information improves syntactic parsing, the reverse is also true: syntactic analysis improves MWE identification. \citet{Greenetal13} successfully tuned a parser for MWE identification, \citet{Weller&Heid2010} and \citet{martens&vandeghinste2010} showed that using parsed corpora for MWE identification is beneficial. These findings led \citet{Seretan2013} to propose that neither accommodating the grammar with MWE information, nor recognizing MWEs in raw text as a help to parsing are appropriate ways of dealing with the issue of MWEs in syntactic parsing because neither approach takes advantage of the fact that MWE information and syntactic analysis are mutually informative. She proposes instead to have a MWE lexicon and to deal with potential MWEs during parsing.\\

\subsection{Advantages and caveats of the three approaches}
\tab All of these researchers have shown the importance of MWEs for syntactic parsing but all of the approaches presented have caveats. Research on HPSG seems to have found the most sophisticated methods of dealing with MWEs but parsing with precision grammars is known to be much less robust \citep{Zhang08robustparsing} (i.e. it fails to parse many more sentences) than parsing with data-induced grammars which make it a not ideal solution for practical parsing. As far as other solutions are concerned, they are often very much limited by the type of MWEs that have been dealt with. All other solutions presented as a matter of fact concentrate on a few types of MWEs. However, as argued by \citet{Kim2008}, because of the different but interrelated properties of MWEs, it is neither appropriate to try and generalize from MWEs and find a single representation which works for all types, nor is it appropriate to deal with each MWE type at a time. An approach for improving syntactic parsing on all MWE types is still lacking and previous approaches leave the question of whether the results can be reproduced with different types of MWEs unanswered. No approach so far has tried to determine if dealing with different types of MWEs in the same way can lead to different results. A lot of work is therefore still needed if we want to take advantage of the full potential that MWE information has to bring to syntactic parsing.

\section{CCG Parsing}
\label{CCGparsing}
\tab  It was mentioned in Section \ref{CCG} that CCG has computational and linguistic properties highly valued in the NLP community. The wish to build efficient parsers with this formalism therefore grew naturally. Recent research has developed efficient tools for parsing with CCG and CCG parsing is increasingly used in NLP applications. It is, to name just a couple of examples, used for question parsing \citep{ClarkSC04} and semantic parsing. \\
\tab The first efficient statistical model was the generative model built by \citet{Hockenmaier2002} and extended by \citet{ClarkCurran2007} to a discriminative model. The first step made by \citet{Hockenmaier2003} was to translate the PTB into a Treebank with CCG derivations, called CCGbank. Both models perform close to state-of-the-art although with simpler statistical models which is argued by the authors to be the result of having a more expressive grammar than the PCFGs used by state-of-the-art parsers.\\
\tab The traditional parsing accuracy metric PARSEVAL has been argued (for example  by \citet{clark2002evaluating}) to be too harsh on CCG derivation trees because they are always binary, as opposed to PTB-style trees which can have flat constructions with more than one branching node. This binary nature of CCG trees make them prone to having more errors. Consequently evaluation of dependencies has generally been preferred for CCG parsing.  For this type of evaluation, labelled and/or unlabelled precision and recall as well as F-measure are computed by comparing dependencies extracted from the output parse trees with dependencies extracted from gold standard trees. As further argued by \citet{clark2002evaluating}, it also makes sense to use dependencies to evaluate CCG parsing since one of the advantages of CCG over other formalisms is precisely its treatment of long-range dependencies.

\section{Research questions and objectives}
\label{relworkccl}

\tab  In Section \ref{syntacticparsing}, it was said that working on syntactic parsing was important for many NLP applications and it was said that one of the promising directions was to try and improve it as a joint task with another NLP task. In Section \ref{syntacticparsingandMWES}, it was said that MWE identification information improves syntactic parsing although current approaches to doing so leave a lot of room for improvement. Trying to improve syntactic parsing with MWE information therefore looks like a promising avenue of research. In Section \ref{syntacticparsing}, advantages of working with a lexicalized grammar formalism for syntactic parsing have been put forward and in Section \ref{CCGparsing}, it was said that parsing with the strongly lexicalized grammar formalism CCG performs almost state-of-the-art. CCG parsing therefore looks like an ideal framework for improving the current approaches to improving syntactic parsing with MWE information. In turn, parsing with a strongly lexicalized grammar formalism is likely to benefit from the addition of lexical information.\\
\tab Very little attention has however been given to MWEs in CCG parsing. \citet{ConstableCurran} modified CCGbank to have a better representation of verb particle constructions but did not report any parsing accuracy improvement. No work has tried to establish whether CCG parsing accuracy could be improved by adding information about MWEs which is what we intend to do in this thesis. Our aim is twofold: we want to find out whether or not MWE information can improve CCG parsing and we want to improve current approaches to adding MWE information to syntactic parsing by avoiding to focus on a very restricted set of MWE types. We wish instead to find out if using methods that have been used for a restricted set of types of MWEs can be extended to different types of MWEs. More details on how we intend to do so will be given in the next chapter. 

\chapter{Methodology}
\label{methodology}
\section{Introduction}
\tab As was made clear in the previous chapter, the objectives of this thesis are to test whether or not information about MWEs can improve CCG parsing as well as to improve previous approaches to adding MWE information to syntactic parsing by avoiding to focus on a restricted set of MWE types and instead trying to find out whether different types of MWE lead to different results. This chapter describes the methodology adopted in this thesis. Section \ref{app} describes the general approach taken and the way in which it relates to previous approaches to integrating MWE information to syntactic parsing. Section \ref{parsingmodel} describes the parsing model upon which we will base the study and which will serve as baseline for the thesis. Section \ref{data} describes the data used. Section \ref{extension} describes the implementation of the changes to the parsing model we propose and Section \ref{trainparse} describes the way in which we train the updated model and parse the test data. Section \ref{eval} describes the evaluation schemes and the way in which they will be used to answer the research questions. The whole pipeline is summed up in Section \ref{pipeline} and we conclude on the contributions of this part in Section \ref{methccl}

\section{Approach}
\label{app}
\subsection{Type of approach}
\label{choiceofapp}
\tab In Section \ref{syntacticparsingandMWES}, three different approaches to adding MWE information to a syntactic parsing model have been described. One possibility is to directly change lexical entries in the lexicon. Another is to recognize MWEs in raw text and collapse MWE units to one token, which can be done to both training and test data for use with a parsing model. A last approach is to build a dictionary of MWEs and change the parsing algorithm so that it takes decisions about MWE representation during parsing. The first approach was proposed for precision grammars and is not easily applicable to a data-induced grammar. The third approach is more sophisticated than the second and has the advantage that it allows to deal with discontiguous MWEs which the second does not, or at least not straightforwardly. The second approach is however already a solid baseline for other formalisms which we propose to implement for CCG parsing in this thesis. Attempting the more sophisticated third approach is beyond the scope of this thesis and will be left to future research.
\subsection{Description of approach}
\label{appdescr}
\begin{figure}[H]
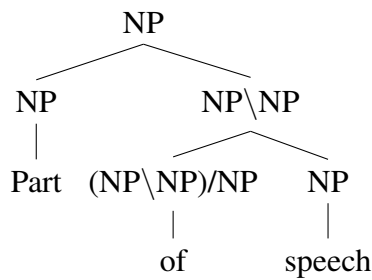

	\Tree [.NP [.NP Part ] [.NP\textbackslash NP [.(NP\textbackslash NP)/NP of ] [.NP speech ] ] ]
	\caption{Traditional tree for Part of speech\label{fig:pos}}
\end{figure}

\begin{figure}[H]
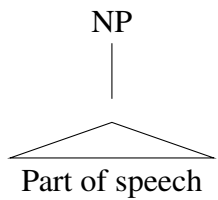

	\Tree [.NP \qroof{Part of speech}. ]
	\caption{Tree for Part of Speech where tokens are grouped \label{fig:pos2}}
\end{figure}

\tab In Section \ref{changingdata}, two different versions of the `changing data' approach have been described. In the first, manual annotation is used to create two versions of the treebank. This left questions unanswered: whether or not the approach can work with automatic recognition and whether or not the approach can work with different types of MWEs. As also said, \citet{Constant2012} have shown that with an automatic recognizer, the parsing accuracy improvement is not as dramatic as predicted by \citet{nivre2004} using gold standard MWE recognition. We will be conducting the type of approach described by \citet{nivre2004} using an automatic recognizer to answer the first of these questions in the context of CCG parsing. We will also experiment with the recognizer by using different versions of it to answer the second of these two questions. This approach is especially interesting in that, as has been shown in \citet{Schneider2013} who attempted a comprehensive annotation of MWEs in a corpus, even manual annotation of MWEs is a difficult task and making automatic MWE recognition an experimental part of the pipeline could lead to interesting results.\\ 
\tab This approach therefore involves changing both training and test data. Changing the training data can help the parsing model learn more sensible representations of language. For example in the tree for part of speech, `of speech' is considered a modifier of `part' as in Figure \ref{fig:pos} which does not make much sense. Instead, grouping the three lexical items as in Figure \ref{fig:pos2} gives a better representation of this group of words. 

\tab Changing the test data by for example collapsing the three lexical items `part', `of' and `speech' to one token `part+of+speech' can help the parser making sensible decisions locally by telling it to consider the three words as one. For example, if this token is followed by a coordinator, the parser knows that coordinating one of the units is not a possibility. In the partial sentence `it gives part+of+speech and lemma information', the parser cannot coordinate `speech' with `lemma' which would be a possibility otherwise. Collapsing MWEs in training and test data lead to two different effects of adding MWE information to the syntactic parsing pipeline and it would be best if we could differentiate both in the experiments. The first type of effect will be called the training effect and the second the parsing effect for convenience. \\
\tab The qualitative analysis of output proposed by \citet{korkontzelos2010} would add a lot of insight to this thesis but will not be conducted for reasons of time and space. As a matter of fact, \citet{korkontzelos2010} built a taxonomy of output differences and used it to quantify each change in the output data but their taxonomy is not readily usable for another syntactic model. In addition, they make extensive use of the information of whether or not a word is given an analysis in their taxonomy. The information that a word is parsed or not is specific to shallow parsing. When using deep parsing, if a word is not parsed, the sentence parse fails. We therefore do not have this type of information. Building a new taxonomy would take us too far afield but we defer it to future work.

\section{Parsing model}
\label{parsingmodel}
\tab As was said in Section \ref{CCGparsing}, both generative and discriminative models exist for parsing with CCG. There are different generative models with different properties. We chose to use StatOpenCCG, developed by \citet{Christodoulopoulos2008} and recently further expanded by \citet{Deoskar} because of its ease of use, flexibility and fast training. The expansion of the parser by \citet{Deoskar} is particularly well suited to our purposes: it was extended so that it works better on unknown lexical items. Collapsing lexical items will increase the sparsity of data and being able to deal with unknown data is therefore a concern for our approach. More particularly, the model proposed by \citet{Christodoulopoulos2008} and \citet{Deoskar} is based on one of \citet{Hockenmaier2003}'s models called LexCat and which conditions probabilities on lexical categories. \citet{Deoskar} make use of this LexCat model instead of the fully lexicalized model which conditions it on words precisely so that the parser is better equipped to deal with unseen lexical items. They introduce a smoothed lexicon to deal with these. They POS-tag the test data in a pre-processing stage and use POS-information to determine the lexical categories of words by using probabilities of lexical categories that appear with each POS-tag of unseen word in the seen data. Because, as mentioned in Section \ref{korkontz}, \citet{korkontzelos2010} have shown that POS-tags assigned automatically to MWEs were useful when parsing, the LexCat model therefore looks ideal for our purposes. We follow \citet{Deoskar} in using the C\&C tools \citep{cctools} to POS-tag our test data so as to have a model that is comparable with theirs.

\section{Data}
\label{data}
\tab All statistical CCG parsing models are based on CCGbank, which, as mentioned in Section \ref{CCGparsing} is a translation of the PTB into CCG derivations built by \citet{Hockenmaier2003} with the objective of building parsing models for CCG. Sections 01-22 are generally used for training, section 00 for development and section 23 for testing. We will follow the tradition so as to have comparable results with previous models.

\section{Extending the parsing model with MWE information}
\label{extension}

\tab As explained in Section \ref{app}, the objective is first to recognize MWEs in the unlabeled version of CCGbank and then to collapse MWEs to one lexical item in the annotated version of the Treebank and in the unlabeled test data. The MWE recognition part is described in Section \ref{mwerec} and the CCGbank conversion is described in Section \ref{ccgconv}. 

\subsection{Recognizing MWEs}
\label{mwerec}
\tab For MWE recognition, we used an existing tool developed by \citet{Finlayson2011}. It is a flexible library which offers many tools useful to our purposes. It can be used to build an index of MWEs with information about their probability. It can also be used with a default index which contains all the MWEs and inflections extracted from Wordnet 3.0 and Semcor 1.6 and statistics for each MWE. There are three different tools of interest to us. Simple detectors detect MWEs in text. There is a detector to find Proper Nouns, one to find all types of MWEs, one that just finds stop words, etc. These simple detectors can also be combined to form a complex detector. There are filters which filter the results of detectors. One for example only accepts MWEs that are continuous, one throws out MWEs which have a score under a certain threshold, one only keeps MWEs under a certain length. The last tool we need is called a `resolver' and it resolves conflicts when lexical items are assigned to more than one MWE. Different resolvers resolve conflicts in different ways: one picks the leftmost MWE, another picks the longest matching MWE.

An example of resolver receives for example the following sentence as input:
\begin{itemize}
	\item{Mr. Spoon said the plan is not an attempt to shore up a decline in ad pages in the first nine months of 1989 ; Newsweek 's ad pages totaled 1,620 , a drop of 3.2 \% from last year , according to Publishers Information Bureau .}
	\end{itemize}
	And gives the following output:
	\begin{itemize}
		\item{mr.\_spoon, shore\_up, according\_to, publishers\_information\_bureau}
	\end{itemize}

	\tab We will need resolvers and will always need to work with a filter that only keeps continuous MWEs but we can experiment with the different resolver types, we can add filters and can experiment with the different types of detectors. This library therefore serves our purposes perfectly since it leaves quite a lot of room for experiments. Experiments are described in Chapter 4. 

	\subsection{Modifying CCGbank}
	\label{ccgconv}
	\tab This section describes the bulk of our work. The idea is to use the list of MWEs obtained from the recognizer and collapse the MWE units in the tree. We worked on trees and dependencies from the trees separately and I will describe each collapsing algorithm in turn.
	\subsubsection{Collapsing MWE units in a CCG parse tree}
	\label{treecoll}
\begin{figure}[H]
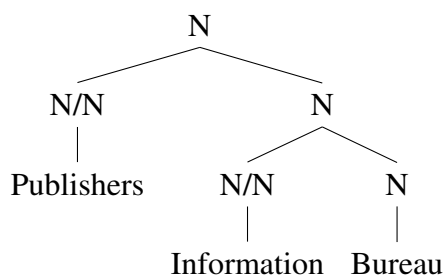

			\Tree [.N [.N/N Publishers ] [.N [.N/N Information ] [.N Bureau ] ] ]
			\caption{Original subtree\label{fig:orst}}
		\end{figure}

\begin{figure}[H]
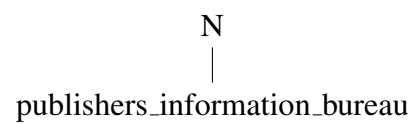

			\Tree [.N publishers\_information\_bureau ]
			\caption{Collapsed subtree\label{fig:cst}}
		\end{figure}

	\tab The algorithm takes a tree and a list of MWEs. It iterates through the list of MWEs, finds them in the tree and collapses MWE units only if they form a constituent in the tree, otherwise the MWE is discarded. The algorithm first looks through the tree leaves to find the index position of each unit of each MWE. Based on these index positions, it finds the tree position that dominates all the units of a MWE. If this tree position dominates all and only the MWE units, the MWE units are siblings in the tree and they are collapsed to one lexical item. The collapsed MWE is assigned the label of the dominating node as a category.\\
	\tab For example, given the MWE publishers\_information\_bureau and the subtree in Figure \ref{fig:orst}, the algorithm returns the subtree in Figure \ref{fig:cst}.

	\begin{multicols}{2}
		\columnbreak
	\end{multicols}

	\tab Pseudocode for the algorithm is given in Algorithm \ref{alg1}. Let MWElist be the list of MWEs for one sentence, let tokens be the MWE units and let leaves be the tree leaves of the sentence. The `matching' function not only checks if the leaf corresponds to the MWE unit but also if the coming leaves match the remaining MWE units of the MWE. For example, if we have the MWE publishers+information+bureau and we find `publishers' but it is followed by `association' or by `information office' the leaf does not match the MWE token. 
	\begin{algorithm}[H]\fbox{\begin{minipage}[c]{\textwidth}
			\begin{algorithmic}
				\State leaf = leftmost tree leaf
				\ForAll{MWE in MWElist} 
				\ForAll{token in MWE}
				\While{matching leaf not found}
				\State check if leaf matches token
				\State look up next leaf to the right
				\EndWhile
				\State get leaf index
				\EndFor
				\State tree = tree spanning indices
				\If{tree spans only indices}
				\State replace tree child with MWE
				\Else
				\State delete MWE
				\EndIf
				\EndFor
				\caption{Collapsing MWE units in the tree if they are siblings\label{alg1}}
		\end{algorithmic}\end{minipage}} \end{algorithm}

\begin{figure}[H]
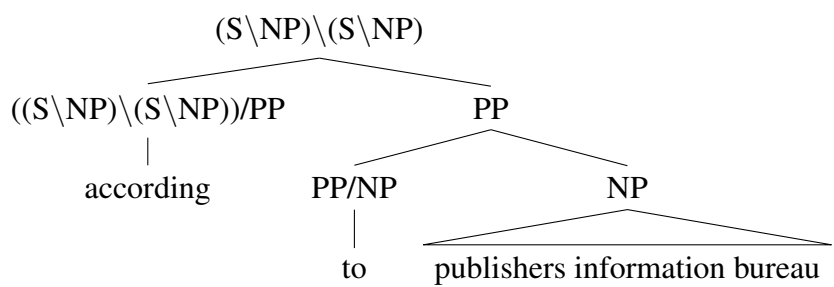

			\Tree [.(S\textbackslash NP)\textbackslash(S\textbackslash NP)
				[.((S\textbackslash NP)\textbackslash(S\textbackslash NP))/PP according ]
				[.PP
					[.PP/NP to ]
			\qroof{publishers information bureau}.NP ] ]
			\caption{Tree with MWE units that are not siblings\label{nst}}
		\end{figure}

\begin{figure}[H]
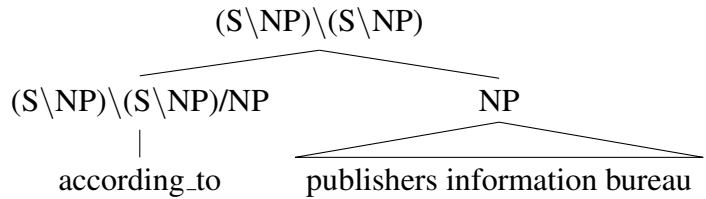

			\Tree [.(S\textbackslash NP)\textbackslash(S\textbackslash NP)
				[.(S\textbackslash NP)\textbackslash(S\textbackslash NP)/NP according\_to ]
			\qroof{publishers information bureau}.NP ]
			\caption{Ideal MWE non-sibling collapsing\label{nsc}}
		\end{figure}

		\tab As mentioned before, the algorithm discards MWEs if they do not form a constituent in the tree. An example of tree in which MWE units (according to) are not siblings in the tree is given in Figure \ref{nst}. The ideal way in which it should be collapsed is given in Figure \ref{nsc} but attempting to find an algorithm which would work for all non-sibling cases was beyond the scope of this thesis. We tried Algorithm \ref{alg1} with a good recognizer, collected statistics and found that 79.5\% of the cases (42,309 out of the 53,208 cases) were siblings in the tree which we considered a good basis for experimentation.

		\subsubsection{Collapsing dependencies}
		\label{depcoll}
\begin{figure}[H]
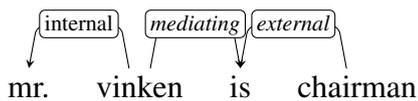

				\begin{dependency}
					\begin{deptext}[column sep=1em]
						mr. \& vinken \& is \& chairman \\
					\end{deptext}
					\depedge{2}{1}{internal}
					\depedge{2}{3}{\textit{mediating}}
					\depedge{4}{3}{\textit{external}}
				\end{dependency}
				\caption{Dependency graph\label{dep4}}
			\end{figure}

\begin{figure}[H]
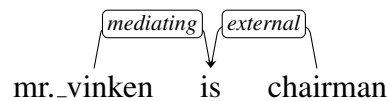

				\begin{dependency}
					\begin{deptext}[column sep=1em]
						mr.\_vinken \& is \& chairman \\
					\end{deptext}
					\depedge{1}{2}{\textit{mediating}}
					\depedge{3}{2}{\textit{external}}
				\end{dependency}
				\caption{Collapsed dependency graph\label{dep3}}
			\end{figure}

		\tab Collapsing the dependencies from the tree is done after having collapsed the MWEs in the tree and requires information about MWE units and their leaf index as well as information about indices of each lexical item in the collapsed tree. If an MWE was discarded in the first algorithm because the units were not siblings, the dependency collapser discards it as well. A dependency is defined as a 6-tuple $<i, j, cat_j, arg_k, word_i, word_j >$ where there is a dependency between $word_i$ and $word_j$ respectively at indices $i$ and $j$. $word_j$ is the functor which has the lexical category $cat_j$ and $word_i$ is the argument and fills the \textit{k}th argument of $word_j$.\\
		\tab Collapsing dependencies involves merging nodes in the graph and changing edges according to the new nodes. When the nodes are collapsed to one token, edges from the original dependency graph fall into three different categories. Let us take the dependency graph in Figure \ref{dep4} as an example in which  `mr. vinken' is a MWE. There are edges between two units of a MWE such as the one between `mr.' and `vinken'. These will be called \textit{internal} edges for convenience. The algorithm deletes them as shown in Figure \ref{dep3}. There are edges between a MWE unit and another word in the sentence such as the edge between `vinken' and `is'. This type of edge will be called a \textit{mediating} edge for convenience. In the collapsed graph, it goes out of the MWE instead of one of its unit as in Figure \ref{dep3}. If the edge was in the other direction, it would similarly come into the MWE instead of one of its units. The edge between `is' and `chairman' does not connect any MWE and does not need changing. It will be called an `\textit{external} edge' for convenience.

		\begin{multicols}{2}
			\columnbreak
		\end{multicols}

		\tab The algorithm deals with each dependency of the tree at a time. If both argument and functor are a MWE unit of the same MWE, the dependency is \textit{internal} and the algorithm deletes it. If $word_i$ is a MWE unit of any MWE, the dependency has a \textit{mediating} edge and the node is replaced with the MWE. If $word_j$ is a MWE unit of any MWE, the dependency also has a \textit{mediating} edge, $word_j$ is replaced with the MWE and $cat_j$ is replaced with the MWE category. In all the remaining dependencies (the \textit{mediating} and \textit{external} edges), the indices are then changed according to the indices in the collapsed version. Pseudocode for the algorithm is given in Algorithm \ref{alg2}. 

		\begin{algorithm}\fbox{\begin{minipage}[b]{\textwidth}
				\begin{algorithmic}[H]
					\ForAll{dependency in dependencies}
					\If{$word_i$ and $word_j$ are MWE units of the same MWE}
					\State dependency has an \textit{internal} edge
					\State delete dependency
					\ElsIf{$word_i$ is a MWE unit of any MWE}
					\State dependency has a \textit{mediating} edge
					\State replace $word_i$ with the MWE
					\ElsIf{$word_j$ is a MWE unit of any MWE}
					\State dependency has a \textit{mediating} edge
					\State replace $word_j$ with the MWE
					\State replace $cat_j$ with the MWE cat
					\EndIf
					\State replace $i$ and $j$ with indices from collapsed leaves
					\EndFor
					\caption{Collapsing dependencies\label{alg2}}
				\end{algorithmic}
		\end{minipage}} 
	\end{algorithm}

\begin{figure}[H]
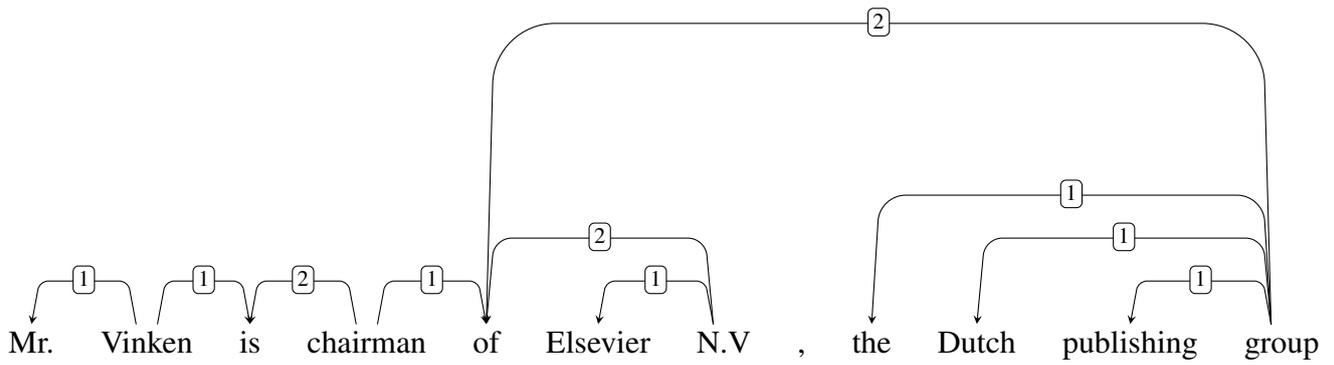

		\hspace{-3em}
		\begin{dependency}
			\begin{deptext}[column sep=1em]
				Mr. \& Vinken \& is \& chairman \& of \& Elsevier \& N.V \& , \& the \& Dutch \& publishing \& group \\
			\end{deptext}
			\depedge{2}{1}{1}
			\depedge{2}{3}{1}
			\depedge{4}{3}{2}
			\depedge{4}{5}{1}
			\depedge{7}{5}{2}
			\depedge{7}{6}{1}
			\depedge{12}{5}{2}
			\depedge{12}{9}{1}
			\depedge{12}{10}{1}
			\depedge{12}{11}{1}
		\end{dependency}
		\caption{Dependency graph\label{dep1}}
	\end{figure}

\begin{figure}[H]
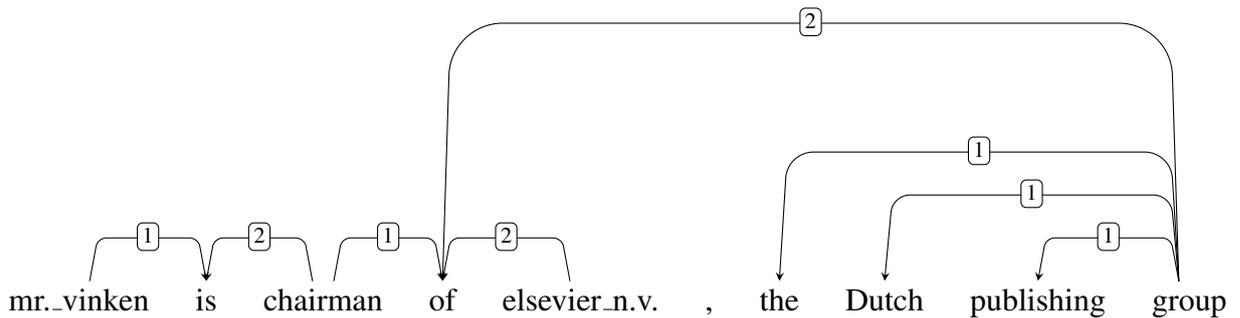

		\hspace{-3em}
		\begin{dependency}
			\begin{deptext}[column sep=1em]
				mr.\_vinken \& is \& chairman \& of \& elsevier\_n.v. \& , \& the \& Dutch \& publishing \& group\\
			\end{deptext}
			\depedge{1}{2}{1}
			\depedge{3}{2}{2}
			\depedge{3}{4}{1}
			\depedge{5}{4}{2}
			\depedge{10}{4}{2}
			\depedge{10}{7}{1}
			\depedge{10}{8}{1}
			\depedge{10}{9}{1}
		\end{dependency}
		\caption{Collapsed dependency graph\label{dep2}}
	\end{figure}

	\tab For the dependency graph in Figure \ref{dep1}, and given the MWEs mr.\_vinken and elsevier\_n.v. the algorithm returns the collapsed version of the graph which is given in Figure \ref{dep2}. The edges labels in the dependency graphs represent $arg_k$.

	\tab The downside of this algorithm is that it can create cyclic dependencies between lexical items, i.e. two nodes are connected by two edges going in the opposite direction. We tested the algorithm on CCGbank and found that in practice this is not a major issue: only 7 cyclic dependencies were created in the \textasciitilde 48.000 sentences.\\

	\section{Training and parsing}
	\label{trainparse}
	\tab To make sure we were working on improving a model that already works well, we first attempted to reproduce the results obtained by \citet{Deoskar} on the non-collapsed data. We used the same parameters to train and parse and obtained similar results (around 87\% of correct lexical categories). This model will serve as our baseline and will henceforth be called model\subscript{A} for convenience.\\
	\tab For each experiment, we run the MWE recognizer on an unlabeled version of CCGbank\footnote{We created an unlabeled version of CCGbank from the tree leaves to make sure the data is compatible with the trees we work with when collapsing.}. We then apply the cascaded algorithms \ref{alg1} and \ref{alg2} to every sentence from CCGbank. We split the data in the traditional way, as explained in \ref{data}, train the model and parse the test file, an unlabeled version of the collapsed test data created from the collapsed trees leaves. We will call the collapsed treebank CCGbank\subscript{B} and the model trained on it model\subscript{B} for convenience.\\
	\tab Using the unlabeled version of the collapsed test raises methodological issues for evaluating, as will be explained in Section \ref{evalB}, because it provides model\subscript{B} with information obtained when collapsing the data, i.e. it implicitly tells it which MWEs are siblings and should be collapsed and therefore makes the parsing task semi-automatic. There is however no straightforward way of detecting siblings in unlabeled data. In addition, the ultimate goal is to collapse all MWEs in the data, regardless of whether or not they are siblings. It was mentioned in Section \ref{ccgconv} that finding such an algorithm involves difficult conceptual and technical issues and is beyond the scope of this thesis but if we had such an algorithm, the problem mentioned would not arise because collapsing unlabeled data would be akin to collapsing annotated data and then converting them to an unlabeled version. Nevertheless, we choose to use this imperfect setting and we will discuss this issue and the solutions we found to circumvent it in Section \ref{evalschemes}.

	\section{Evaluation}
	\label{eval}
	\tab After having trained our model\subscript{A} and model\subscript{B}s, we are equipped to answer the research questions from Section \ref{relworkccl}, namely, we just need to evaluate them in different ways. In this Section, I first describe the evaluation metrics we will be using and then I describe the different evaluation schemes we implemented to answer each of the research questions.
	\subsection{Evaluation metrics}
	\label{evalmetrics}
	\tab As was mentioned in Section \ref{CCGparsing}, evaluating dependencies has often been preferred to evaluating bracketing for CCG parsing. Alternatively, supertags are sometimes evaluated. In our case, we are not especially interested in supertags. We have changed some of them in the gold standard as explained in Section \ref{treecoll} but our approach changes the structure of the tree in the gold standard and it is in this structure that we believe we have proposed a better data representation. We therefore decided to follow the first mentioned tendency to use the dependency evaluation scheme in this thesis. More specifically, dependencies of output parses are compared against gold standard and labelled and unlabelled precision and recall are computed. Precision is computed as in Equation \ref{prec}, Recall is computed as in Equation \ref{recall} and a harmonic F-measure can be obtained with the formula in Equation \ref{F}. Because we have not attempted to work on dependency labels in the dependency collapsing algorithm, we will exclusively be making use of scores on unlabelled dependencies.
	\begin{equation} \label{prec}
		Precision = \frac{\#correct\:dependencies}{\#attempted\:dependencies}
	\end{equation}
	\begin{equation}\label{recall}
		Recall = \frac{\#correct\:dependencies}{\#dependencies\:in\:gold\:standard}
	\end{equation}
	\begin{equation} \label{F}
		F\beta = \frac{(\beta^2 + 1)PR}{\beta^2P+R}
	\end{equation}\\
	\tab We will always be using the harmonic F measure with $\beta$ = 1 resulting in the simpler formula in Equation \ref{F1}.\\
	\begin{equation} \label{F1}
		F1 = \frac{2PR}{P+R}
	\end{equation}\\
	\subsection{Evaluation schemes}
	\label{evalschemes}
	\tab In order to evaluate our models, we can extract dependencies from the parsed files and compare them with the gold standard. This way, we can compare different models and answer the research questions. Because we changed the gold standard as compared to model\subscript{A} (as defined in the previous section) however, the results obtained from comparing our parsed files with our gold standard are not directly comparable with the results obtained when applying the same evaluation scheme to model\subscript{A} and we cannot directly compare model\subscript{A} with our model\subscript{B}s (as defined in the previous section) which is essential in answering our research questions. Instead, we have to change the data of one of the models so as to compare each of the models with the same gold standard. Because we assume that we have created sensible gold standard with our collapsing method, we mainly used gold standard\subscript{B} for evaluation and call that Eval vs Gold\subscript{B} for convenience. \\
	\tab As was said in Section \ref{appdescr}, changing training and test data can lead to two different effects which can lead to an improved parsing accuracy, i.e. training (the parser learns useful information during training) and parsing effects (the trained parser is helped in its decisions by MWE information) which we said we would like to differentiate. This can be achieved by conducting different experiments with our existing models. We can assess training effect by testing whether or not model\subscript{B} can beat model\subscript{A} when evaluated on the same gold standard. We can assess parsing effect by testing whether or not model\subscript{A} can beat itself when given collapsed test data. We discuss these evaluation schemes in Section \ref{evalB} \\
	\tab If there is training and/or parsing effect, we can assume that automatic recognition of MWEs can improve syntactic parsing and hence answer the first of our research questions. We can verify this by testing whether or not we can use information from model\subscript{B} to beat model\subscript{A} on gold standard\subscript{A}. We implemented a second evaluation scheme where we combine information from output from \modelA and output from \modelB (henceforth called model combination) to test this which we discuss in Section \ref{evalA}. \\
	\tab As was mentioned in Section \ref{relworkccl}, we not only want to know whether or not information about MWEs can help CCG parsing but we are also interested in finding out whether or not different types of MWEs impact parsing accuracy in different ways. As will be explained in Chapter \ref{exp}, we created different versions of CCGbank\subscript{B} and different versions of model\subscript{B} with these. Because we created different gold standard for each of these models, they cannot directly be compared. Instead, comparing how different model\subscript{B}s can improve model\subscript{A} is possible by comparing their combination with it against gold standard\subscript{A}. Again then we can use model combination and combine information from output from \modelA with information from output from \modelB. We compare this combined model output against gold standard\subscript{A} and compare the results when combining \modelA with different versions of \modelB. We discuss how this can be achieved in Section \ref{evalB2}. \\
	\subsubsection{Assessing training and parsing effects}
	\label{evalB}
	\tab Modifying the output from model\subscript{A} so that it is comparable with gold standard from model\subscript{B} is straightforward: we just need to apply the collapsing algorithms to the output from model\subscript{A} with the MWEs found in the test data. We can also test \modelA on data collapsed before parsing.
	\paragraph*{parsing effect}
	Testing whether there is a parsing effect can be done by testing whether or not model\subscript{A} can perform better on test data collapsed before parsing than on test data collapsed after parsing. We can therefore test model\subscript{A} on the collapsed test data and see if it improves on model\subscript{A} tested on unchanged data. We conducted this evaluation. In this evaluation however, the caveat mentioned in Section \ref{trainparse} that we are using information from the gold standard in the test data, i.e. we know which MWEs are siblings in the test data is problematic. This introduces an artefact which makes the results somewhat difficult to interpret: the collapsing before parsing method has sibling information which the collapsing after parsing method does not. There can be parsing and sibling effects and the two cannot be decoupled. A way to circumvent this problem is to collapse MWEs regardless of their sibling status (i.e. treat all detected MWEs as if they were siblings) and compare the model when we collapse before parsing with the model when we collapse after parsing. Collapsing all MWEs in unlabeled test data is straightforward. It was however mentioned in Section \ref{treecoll} that we do not have an algorithm to collapse trees so we cannot collapse the output parse trees. However, since we are working only with dependencies for evaluation, it is possible to collapse all MWEs in all the dependencies of the sentence. The problem with this evaluation is that the output cannot perform well on gold standard\subscript{B} because it is not tokenized in the same way and we treat dependencies wrongly tokenized as errors. Both collapsing before and collapsing after parsing however suffer from the same problem and the comparison between the two is fair so we will be using this evaluation to test whether or not there can be a parsing effect.
	\paragraph*{training effect}
	Testing whether there is a training effect consists in comparing the results of model\subscript{A} on collapsed data with the results of model\subscript{B} on collapsed data. In this evaluation, the caveat that we are using information from the gold standard in the test data can also be considered problematic because model \subscript{B} was trained on data with information about siblings which model\subscript{A} was not and is possibly better suited to deal with unseen data with sibling information. We therefore both tested both models on data where only siblings are collapsed (called `gold test' for convenience) and on data where all MWEs are collapsed (called `fully collapsed test' for convenience). Again, the problem with this evaluation is that the output cannot perform well on gold standard\subscript{B} because it is not tokenized in the same way. Again however both models suffer from the same problem and the comparison between the two is fair so this evaluation can be used to test whether or not there is a training effect. \\
	\subsubsection{Verifying whether or not automatic recognition of MWEs can improve CCG parsing on the original gold standard}
	\label{evalA}
\begin{figure}[H]
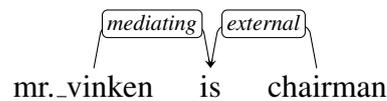

		\center
		\begin{dependency}
			\begin{deptext}[column sep=1em]
				mr.\_vinken \& is \& chairman \\
			\end{deptext}
			\depedge{1}{2}{\textit{mediating}}
			\depedge{3}{2}{\textit{external}}
		\end{dependency}
		\caption{Collapsed dependency graph\label{dep5}}
	\end{figure}

	\tab Results which will be discussed in Chapter 4 seem to indicate that there is both a training and a parsing effect and that model\subscript{B} performs better than model\subscript{A} on some dependencies. Our findings support the claim that automatic recognition of MWEs can improve CCG parsing. These results however led us to want to check whether or not model\subscript{B} can and improve the score on the standard evaluation benchmark, i.e. on gold standard A. This evaluation scheme will be called Eval vs Gold\subscript{A} for convenience. This involves `decollapsing' the output from model\subscript{B} and splitting MWEs back into their units. However, by collapsing the data, we have lost information about some dependencies in the sentence. We have no \textit{internal} edges (edges between MWE units of the same MWE) and when there is a \textit{mediating} edge (edges between MWE units of any MWE and other words in the sentence) we do not know which MWE unit of the MWE it should go to or come from. In Figure \ref{dep3} reproduced in Figure \ref{dep5} for convenience, we do not know whether the label between is and `mr.\_vinken' should come from `mr' or `vinken'.

	\tab \textit{External} edges can be taken directly from model\subscript{B}. \textit{Internal} edges do not exist in model\subscript{B}. Hence we propose to take them from model\subscript{A} and implement a `model combination' algorithm which combines dependencies from \modelA with dependencies from model\subscript{B}. For \textit{mediating} edges, there are different possibilities. Either we can take them from model\subscript{A} and therefore only test whether or not model\subscript{B} performs better than model\subscript{A} on \textit{external} edges. This was implemented as one of the evaluation schemes which will be called Eval vs Gold\subscript{B} `medFromA' for convenience. If we want to test model\subscript{B} on \textit{mediating} edges however, the model combining algorithm will have to choose the node it comes from or goes to: in our example, it should either be `mr.' or `vinken'. The following two other evaluation schemes were implemented, one in which the rightmost node is chosen as incoming or outcoming node for \textit{mediating} edges from model\subscript{B}, which I will henceforth call the `rightmostMed' Eval vs Gold\subscript{B} scheme, one in which the leftmost node is chosen which I will henceforth call the `leftmostMed' scheme. In order for the model combining algorithm to work, we need to recover information about MWEs so that we know for each word in the dependency whether or not we are dealing with a unit of a MWE and hence to know if we are dealing with an \textit{internal}, \textit{external} or \textit{mediating} edge. This can easily be done because MWE units are marked as such in the unlabeled data\footnote{They are joined by a `+' symbol.}.\\
	\tab When these models are combined in these three different ways, we have a new combined model that we can compare with \modelA on gold standard\subscript{A}. In this case again however using `gold test' data is problematic. As a matter of fact, if we use output\subscript{B} as obtained after parsing `gold test' data, we are using information obtained during the conversion of the gold standard and we are using a parsing pipeline which is not fully automatic. In order to make sure that we can beat \modelA in a fully automatic way, we can use parses of \modelB tested on the test data which were collapsed as though all MWEs were siblings as described in Section \ref{evalB}, which will henceforth be called the `fully collapsed test' data, and combine them in the same three ways as described above.

	\subsubsection{Testing whether or not different MWE types impact the results differently}
	\label{evalB2}
	\tab As was said before, we used different MWE recognizers to create different versions of CCGbank\subscript{B} and hence different versions of model\subscript{B}. As also said, because we created different gold standard for each, results from different models are not directly comparable. We can however convert output using the model combination algorithm described in Section \ref{evalB} and test each model against gold standard A. In this way, different versions of model\subscript{B} can be compared.

	\section{Summary of the pipeline}
	\label{pipeline}
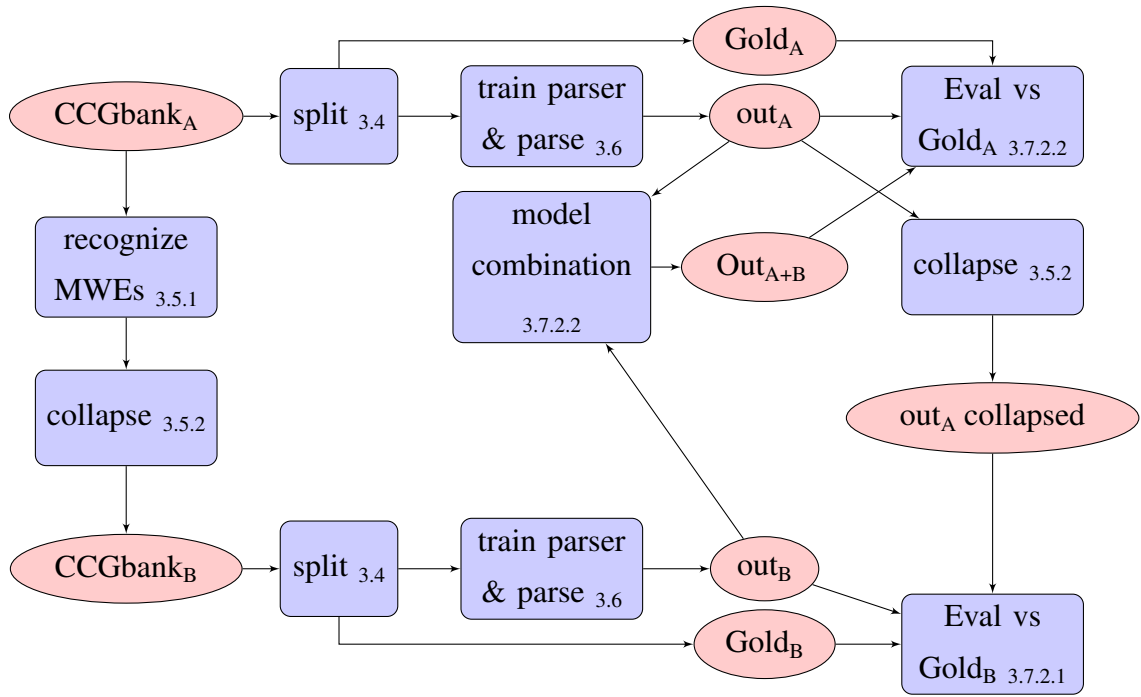
\begin{figure}[H]
		\begin{tikzpicture}[node distance = 2cm, auto]

			\node[cloud] (init) {CCGbank\subscript{A}};
			\node[block, right of=init, node distance=2.8cm, text width = 3em] (split) {split \subscript{\ref{data}}};
			\node[block, right of=split, node distance=2.8cm] (traintest) {train parser \& parse \subscript{\ref{trainparse}}};
			\node[cloud, right of=traintest, node distance=2.8cm] (outA){out\subscript{A}};
			\node[cloud, above of=outA, node distance =1cm](GSA){Gold\subscript{A}};
			\node[block, right of=outA, node distance =3cm](ResA){Eval vs Gold\subscript{A} \subscript{\ref{evalA}}};
			\node[block, below of=ResA, node distance =2cm](collA){collapse \subscript{\ref{ccgconv}}};
			\node[cloud, below of=outA, node distance =2cm, text width = 3em](comb){Out\subscript{A+B}};
			\node[cloud, below of=collA, node distance =2cm](outcollA){out\subscript{A} collapsed};
			\node[block, below of=init, node distance=2cm] (rec) {recognize MWEs \subscript{\ref{mwerec}}};
			\node[block, below of=rec, node distance=2cm] (collapse) {collapse \subscript{\ref{ccgconv}}};
			\node[cloud, below of=collapse, node distance=2cm] (B) {CCGbank\subscript{B}};
			\node[block, right of=B, node distance=2.8cm, text width = 3em] (splitB) {split \subscript{\ref{data}}};
			\node[block, right of=splitB, node distance=2.8cm] (traintestB) {train parser \& parse \subscript{\ref{trainparse}}};
			\node[cloud, right of=traintestB, node distance=2.8cm] (outB){out\subscript{B}};
			\node[cloud, below of=outB, node distance =1cm](GSB){Gold\subscript{B}};
			\node[block, below of=traintest, node distance =2cm, text width = 5.5em] (combine) {model combination \subscript{\ref{evalA}}};
			\node[block, right of=GSB, node distance =3cm](ResB){Eval vs Gold\subscript{B} \subscript{\ref{evalB}}};

			\path [line] (init) -- (split);
			\path [line] (split) -- (traintest);
			\path [line] (traintest) -- (outA);
			\path [line] (split) |-(GSA);
			\path [line] (outA) -- (ResA);
			\path [line] (outA) [bend left=45]-- (combine);
			\path [line] (comb)[bend right=45] -- (ResA);
			\path [line] (outA) [bend right=45] -- (collA) ;
			\path [line] (collA) -- (outcollA);
			\path [line] (outcollA) -- (ResB);
			\path [line] (GSA) -| (ResA);
			\path [line] (init) -- (rec);
			\path [line] (rec) -- (collapse);
			\path [line] (collapse) -- (B);
			\path [line] (B) -- (splitB);
			\path [line] (splitB) -- (traintestB);
			\path [line] (traintestB) -- (outB);
			\path [line] (outB) [bend right=45] -- (ResB);
			\path [line] (outB) -- (combine);
			\path [line] (combine) -- (comb);
			\path [line] (GSB) -- (ResB);
			\path [line] (splitB) |-(GSB);
		\end{tikzpicture}
		\caption{Pipeline of an experiment on one version of one application of MWE recognition to the parsing pipeline with all the evaluation schemes that can be applied to it. The collapsing before parsing of model A (see Section \ref{evalB}) is omitted for clarity and given in Figure \ref{pipel2}. \label{pipel}}
	\end{figure}

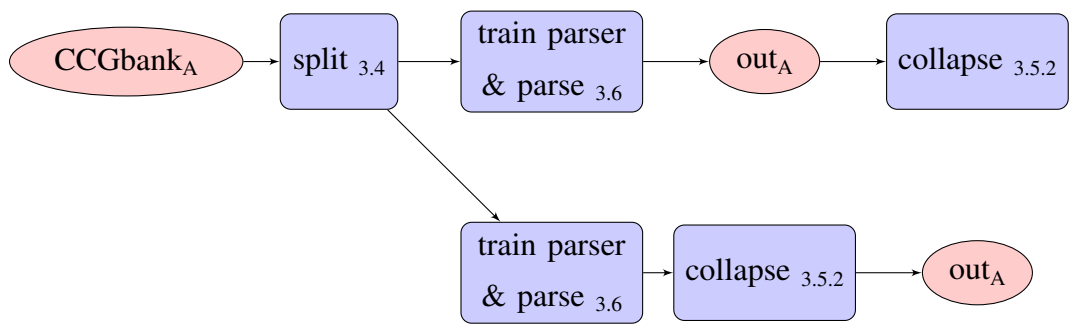
\begin{figure}[H]
		\begin{tikzpicture}[node distance = 2cm, auto]

			\node[cloud] (init) {CCGbank\subscript{A}};
			\node[block, right of=init, node distance=2.8cm, text width = 3em] (split) {split \subscript{\ref{data}}};
			\node[block, right of=split, node distance=2.8cm] (traintest) {train parser \& parse \subscript{\ref{trainparse}}};
			\node[cloud, right of=traintest, node distance=2.8cm] (outA){out\subscript{A}};
			\node[block, right of=outA, node distance=2.8cm] (collapse) {collapse \subscript{\ref{ccgconv}}};
			\node[block, below of=traintest, node distance=2.8cm] (traintestB) {train parser \& parse \subscript{\ref{trainparse}}};
			\node[block, right of=traintestB, node distance=2.8cm] (collapseB) {collapse \subscript{\ref{ccgconv}}};
			\node[cloud, right of=collapseB, node distance=2.8cm] (outB){out\subscript{A}};

			\path [line] (init) -- (split);
			\path [line] (split) -- (traintest);
			\path [line] (traintest) -- (outA);
			\path [line] (outA) -- (collapse);
			\path [line] (split)[bend right=45] -- (traintestB);
			\path [line] (traintestB) -- (collapseB);
			\path [line] (collapseB) -- (outB);

		\end{tikzpicture}
		\caption{Pipeline of the "collapsing before parsing" against the "collapsing after parsing experiment" \label{pipel2}}
	\end{figure}

	\tab The whole pipeline of an experiment is summarized in Figure \ref{pipel}. First CCGbank\subscript{A} is split and the parser is trained and tested. Then MWEs are recognized in CCGbank as described in Section \ref{mwerec} which is subsequently collapsed as described in Section \ref{ccgconv}. Data are split and the parser is trained and the test set is parsed. Output from model\subscript{A} and model\subscript{B} are combined in the three possible ways described in \ref{evalB} so as to allow the evaluation of parts of model\subscript{B} and model\subscript{A} against gold standard\subscript{A} . Output\subscript{A}  is collapsed so as to allow the evaluation model\subscript{B} and model\subscript{A} against gold standard\subscript{B}. The `collapsing before parsing' of \modelA is not included in the Figure but represented in Figure \ref{pipel2}. In this case, test data is collapsed before parsing and the output of \modelA can be compared against gold standard\subscript{B}. This can then be compared to \modelA when the output is collapsed after parsing as shown in the Figure. The Section in which each script is described is given in each script box in both Figures.

	\section{Conclusion}
	\label{methccl}
	\tab In Chapter \ref{relatedwork}, we have motivated our study and identified two research questions: whether or not information about MWEs can improve CCG parsing and whether or not different types of MWEs can influence parsing accuracy in different ways. In this Chapter, we have proposed a methodology for testing this. We have refined the first research questions: what we want to find out is whether or not automatic recognition of MWEs can improve CCG parsing. Additionally, we have separated it into two further research questions: whether we can observe a parsing effect (the parser is helped in its decisions by collapsed data) and/or whether we can observe a training effect (the parser learns something useful). We have proposed to use different MWE recognizers to answer the second question. When defining an algorithm for collapsing the Treebank however, we could not find a straightforward algorithm to collapse MWEs that are not siblings in the tree and decided to settle for an algorithm that only collapses siblings. This led to further complications in the evaluation schemes because it makes it harder to give a fair evaluation of our models. We however found ways to circumvent the problems by implementing different evaluation schemes together with cross-validations. We have tested our methodology using different recognizers and we present the results in the next Chapter.

\chapter{Experiments and results}
\label{exp}

\section{Introduction}

\paragraph*{Overview}
\tab As mentioned before, the objectives of this thesis are first to find out whether or not automatic MWE recognition can be useful to parsing with CCG and second to find out whether or not applying the same approach using different types of MWEs in the MWE recognition stage can lead to different results. In more practical terms, the first question is, can we create a model which has MWE information which model\subscript{A} does not have and which performs better than model\subscript{A}? In Chapter \ref{methodology}, this question was also decomposed into two sub-questions: whether or not there is a training effect from collapsing the data, i.e. the parser learns something useful and whether or not there is a parsing effect, i.e. the parser is helped in its decisions by collapsed data. The second research question is less precise: do we observe differences in results when using different MWE recognizers? What would this tell us about MWE recognition and its interaction with syntactic parsing? In this chapter, I will deal with each of these two questions in turn: first dealing with the general effect of adding MWE information to CCG parsing and second I will deal with observations that can be made when using different recognizers. Before I deal with these questions, I will introduce the different MWE recognizers that we used.
\paragraph*{Significance tests}
We used a one-tailed Randomized Shuffling test with 10,000 iterations to asses statistical significance of our best results. This test is particularly suited to our purposes, it was created for research in Computational Linguistics which assesses precision, recall and F-scores and has the advantage over most other statistical tests that it does not make independence assumptions between the models compared which we should not be making in our case. We used the software created by \citet{sigf06} (slightly modified in order to make it a one-tailed test instead of a two-tailed one) for our tests.

\section{MWE recognition}
\label{expmwerec}
\tab As mentioned in Section \ref{mwerec}, we used three different tools from the MWE recognition library. Detectors which detect MWEs in text, filters which filter through the results of one or more detectors and resolvers which resolve conflicts between MWEs when one word is assigned to more than one MWEs by the detector. We built 5 different MWE recognizers with these three tools. This means that the study is by no means exhaustive, we just tried combinations that were comparable with each other and that intuitively made sense. Building all possible combinations was beyond the scope of the thesis. \\
\tab We used two different resolvers: one which always picks the longest matching MWE (which will be called `Longest'), one which picks the leftmost MWE (which will be called `Leftmost', i.e. the one that started earliest in the sentence). We used 3 different detectors: one which finds all MWEs (which will be called `Exhaustive', one which finds only Proper Nouns (which will be called `ProperNoun') and one which finds only stop words (which will be called `StopWords'. We always used the filter that only keeps continuous MWEs. We used a filter that only keeps an MWE if its units appear more often in the MWE than as a single token which will be called MoreFrequentAsMWE. We used a filter which constrains the length of MWEs to two which will be called `ConstrainLength'. We summarize the recognizers used as well as information collected during the collapsing algorithm in Table \ref{tab:rec}. The numbers in column `ID' will be used throughout this section to denote the recognizer used. Similarly, each \modelB will be denoted by the recognizer which was used to train it as indicated by this number.

\begin{table}[H]
	\footnotesize
	\def\arraystretch{1.20} 
	\begin{tabular}{| l | l | l | l | c | c | c |} \hline
		\textbf{ID}
		& \textbf{detector}
		& \textbf{filter}
		& \textbf{resolver}
		& \textbf{MWE count}
		& \textbf{Sibling count}
		& \textbf{Sibling \%}
		\\ \hline
		1 & Exhaustive & MoreFrequentAsMWE & Longest & 53,208 & 42,309 & 79.51 \\
		2 & Exhaustive & MoreFrequentAsMWE & Leftmost & 51,543 & 21,532 & 41.85 \\
		3 & Proper Nouns &no filter & Longest & 32,583 & 28,068 & 86.14 \\
		4 & Exhaustive & ConstrainLength & Leftmost & 49,587 & 19,984 & 40.30 \\
		5 & Stop words & no filter & Longest & 13,623 & 286 & 2.09 \\
		\hline
	\end{tabular}
	\caption{Description of MWE recognizers used}
	\label{tab:rec}
\end{table}

\section{Can we improve CCG parsing accuracy with automatic MWE recognition?}
\label{q1}
\tab As said in Section \ref{eval}, we implemented different evaluation schemes to answer this question. First we evaluate model\subscript{B} and model\subscript{A} against gold standard\subscript{B} and determine whether there is training and/or parsing effects. Then we check whether we can use model\subscript{B} to improve model\subscript{A} on gold standard\subscript{A} by using model combination with \modelA and model\subscript{B}. I will deal with each of these in turn. We tested all evaluation schemes on all of our versions of model\subscript{B}. Results fluctuate according to the recognizers as will be discussed in Section \ref{q2}. We give general remarks about results and give our best results in this Section as well as their significance level.
\subsection{Can MWE collapsing introduce training and parsing effects?}
\tab I will first try to determine whether or not we can observe a training effect when training model\subscript{B}, then I will try to determine whether or not we can observe a parsing effect when collapsing data before and after parsing with model\subscript{A}. 
\subsubsection{Can MWE collapsing introduce a training effect?}

\tab As described in Section \ref{evalB}, in order to find out whether or not training data on a MWE-informed corpus can lead to an improved accuracy, i.e. leads to training effect, we compared the output of \modelB against the output of \modelA tested on the `gold test' data. 3 out of our 5 model\subscript{B} outperform model\subscript{A} on unlabeled\subscript{B}, although generally by a slight margin. The best results are obtained by model\subscript{B\subscript{3}} and are given in Table \ref{tab:res1}. Model\subscript{B} significantly outperforms \modelA by 0.24\% (p=.006) which supports the hypothesis that there is indeed a training effect.

\begin{table}[H]
	\footnotesize
	\def\arraystretch{1.20} 
	\centering
	\begin{tabular}{| l | l | c | c | c |} \hline
		\textbf{model}
		& \textbf{test data}
		& \textbf{P}
		& \textbf{R}
		& \textbf{F$_1$}
		\\ \hline
		A & gold test &\textbf{84.53}  &84.76 &84.64 \\
		B3 & gold test &84.48  &\textbf{85.28}  &\textbf{84.88} \\
		\hline
	\end{tabular}
	\caption{Precision (P), recall (R), and F$_1$-measure of unlabelled dependencies against gold standard B with recognizer 3 \label{tab:res1}}
\end{table}

\tab As mentioned in the same section however, because using gold test data might bias the results towards model\subscript{B}, we also tested these models on the `fully collapsed test' data. In this case 3 of our 5 model\subscript{B}s outperform \modelA again although by an even slighter margin. The biggest difference in results is obtained with model\subscript{B\subscript{1}} and results are given in Table \ref{tab:res1bis}. Although the margin is smaller, \modelB still significantly outperforms \modelA by 0.15\% (p=.047) which shows that there clearly is a training effect.

\begin{table}[H]
	\footnotesize
	\def\arraystretch{1.20} 
	\centering
	\begin{tabular}{| l | l | c | c | c |} \hline
		\textbf{model}
		& \textbf{test data}
		& \textbf{P}
		& \textbf{R}
		& \textbf{F$_1$}
		\\ \hline
		A & fully collapsed test &\textbf{73.15}  &72.38  & 72.77 \\
		B3 & fully collapsed test &73.08  &\textbf{72.74}  & \textbf{72.92} \\
		\hline
	\end{tabular}
	\caption{Precision (P), recall (R), and F$_1$-measure of unlabelled dependencies against gold standard B with recognizer 1 \label{tab:res1bis}}
\end{table}

\subsubsection{Can MWE collapsing introduce parsing effect?}
\tab As described in Section \ref{evalB}, we compared the output of \modelA when data are collapsed before parsing with \modelA when data are collapsed after parsing. In this case \modelB always outperforms model\subscript{A}. Our best results are shown in Table \ref{tab:res1bisbis} in which \modelB highly significantly outperforms \modelA (p$<$.0001).

\begin{table}[H]
	\footnotesize
	\def\arraystretch{1.20} 
	\centering
	\begin{tabular}{| l | l | l | c | c | c |} \hline
		\textbf{model}
		& \textbf{collapsed}
		& \textbf{P}
		& \textbf{R}
		& \textbf{F$_1$}
		\\ \hline
		A & before parsing &\textbf{83.88}  &\textbf{84.24}  & \textbf{84.06} \\
		A & after parsing &78.92  &79.41  & 79.17  \\
		\hline
	\end{tabular}
	\caption{Precision (P), recall (R), and F$_1$-measure of unlabelled dependencies against gold standard B with recognizer 1 when collapsing before parsing uses gold sibling information and collapsing after parsing collapses only siblings\label{tab:res1bisbis}}
\end{table}

\tab However, as explained in Section \ref{evalB}, these results have a caveat because the `collapsing before parsing' method has gold standard information about siblings which the `collapsing after parsing' method does not. It is interesting despite of this caveat but as we also said in that Section, we have a way of verifying whether or not MWE information is partly responsible for this result. For this verification, we collapse all MWEs both before and after parsing and compare the results. In this case, \modelA when data are collapsed before parsing outperforms \modelA when data are collapsed after parsing only in one of the 5 cases which undermines a little the previous argument about the parsing effects showing that there can also be undesirable effects to collapsing test data. It could however be partly due to the fact that we collapsed non-siblings, which may have triggered errors during parsing. In any case, our best results still show a significant improvement with the `collapsing before parsing' method over the `collapsing after parsing' method . These results are obtained when using recogniser\subscript{3} and are given in Table \ref{tab:res1tris}. Model\subscript{A} collapsed before parsing significantly beats \modelA collapsed after parsing by 0.20\% (p=.008). This indicates that there is a parsing effect.

\begin{table}[H]
	\footnotesize
	\def\arraystretch{1.20} 
	\centering
	\begin{tabular}{| l | l | c | c | c |} \hline
		\textbf{model}
		& \textbf{collapsed}
		& \textbf{P}
		& \textbf{R}
		& \textbf{F$_1$}
		\\ \hline
		A & before parsing &\textbf{79.83} & 79.54  & \textbf{79.69} \\
		A & after parsing &79.38  &\textbf{79.60}  & 79.49 \\
		\hline
	\end{tabular}
	\caption{Precision (P), recall (R), and F$_1$-measure of unlabelled dependencies against gold standard B with recognizer 3 when all MWEs are considered siblings \label{tab:res1tris}}
\end{table}

\subsection{Can we improve the parsing model on the original gold standard?}

\tab As was said in Section \ref{evalA}, because we observed improvements in accuracy from \modelA to \modelB on gold standard\subscript{B} which showed that \modelB seemed to be better than model\subscript{A}, we decided to test whether or not \modelB can improve \modelA on gold standard\subscript{A}. This means testing whether or not \modelB improves on \modelA on \textit{external} edges and/or on \textit{mediating} edges. As a reminder, we decided to test this with three different evaluations testing three different combinations of \modelA and \modelB. \textit{Internal} edges are always taken from \modelA and \textit{external} edges are always taken from model\subscript{B}. \textit{Mediating} edges are taken from A in the `medFromA' evaluation and from B in the 2 other cases. In the `rightmostMed' evaluation, the rightmost MWE unit is always chosen as incoming or outcoming node and in the `leftmostMed' evaluation, it is the leftmost MWE unit that is always taken as incoming or outcoming node. Our best results are given in \ref{tab:res2} in which \modelB only outperforms \modelA in the `medFromA' case by 0.13\% which is not significant (p$>$.05). This seems to show that \modelB may perform better than \modelA on \textit{external} edges but as far as \textit{mediating} edges are concerned, the picture is unclear. If we take the \textit{mediating} edge from B, it seems clearly better to choose the rightmost MWE unit as incoming or outcoming node (which is not surprising since heads of compound nouns are almost always rightbranching) but doing so does not seem to be a big help in parsing accuracy. Model\subscript{B} might perform better than \modelA on \textit{mediating} edges if we had a better mechanism to recover the head word but with our simple method we cannot say whether or not it is the case.

\begin{table}[H]
	\footnotesize
	\def\arraystretch{1.20} 
	\centering
	\begin{tabular}{| l | l | c | c | c |} \hline
		\textbf{model}
		& \textbf{combination type}
		& \textbf{P}
		& \textbf{R}
		& \textbf{F$_1$}
		\\ \hline
		A & & \textbf{85.27} & 85.02 & 85.15\\
		A+B3 & medFromA &84.89  &\textbf{85.68} & \textbf{85.28} \\
		A+B3 & rightmostMed &84.84  &85.46  & 85.15 \\
		A+B3 & leftmostMed &81.43  &82.02  & 81.72 \\
		\hline
	\end{tabular}
	\caption{Precision (P), recall (R), and F$_1$-measure of unlabelled dependencies against gold standard A using recognizer 3\label{tab:res2}}
\end{table}

\tab As said in Section \ref{evalA}, in the evaluation presented here, \modelB is again helped in the parsing decisions by being told which MWEs are siblings and the resulting parsing pipeline is therefore not fully automatic. In order to test whether or not we could improve on \modelA in a fully automatic manner, we tested \modelB on the `fully collapsed test' data which is a version of the test data obtained automatically, i.e. by collapsing all MWEs in the text instead of only the siblings. All MWEs are then parsed as a unit. When we combine the models, we have more MWEs than we should have and consequently, more edges are considered to be \textit{mediating} and \textit{internal} edges and less edges are considered to be \textit{external} edges. Hence, we are led to choose edges from \modelA where \modelA is not expected to perform better than \modelB. When combining both models with the `medFromA' method however, we still outperform \modelA by 0.04\% when using recognizer\subscript{3} showing that \modelB may have learnt something useful although there is no significant evidence for it at this point. \\
\tab Overall then it seems that \modelB can outperform \modelA on different evaluation schemes and we can conclude that \modelB can be better than \modelA at least on some dependencies. It seems that we can improve CCG parsing with automatic MWE recognition although the improvements we have seen are not dramatic and we have failed to show that \modelB can significantly outperform \modelA on gold standard A.
\section{Does using different MWE recognizers impact parsing accuracy differently?}
\label{q2}
\tab As explained in Section \ref{evalB2}, the last experiment we attempted was to test our model on different recognizers, combine the output using the model combination algorithm explained in Section \ref{evalB} and compare it to gold standard\subscript{A}. This provides a way to compare different versions of our model\subscript{B}.\\	
\tab As can be seen in Table \ref{tab:res3}, different MWE recognition methods seem to make a difference in results. There is a significant difference between our best model (based on recognizer\subscript{3}) and our worst model (based on recognizer\subscript{2}) of .26 (p=.01). Some recognizers decrease parsing accuracy while others increase it. It appears from the table that using a leftmost resolver (a resolver that always chooses the leftmost MWE when there is a conflict) has a bad impact on parsing accuracy. Looking at the different models, it is interesting to note that there is a much lower percentage of MWEs that are siblings in the tree and hence a much lower amount of changes made in the treebank. It is interesting to note that the best model is based on a detector that only detects Proper Nouns. This seems to show that they are the best candidates for being treated as words-with-spaces which is not surprising because they are not flexible and never get inflected. For other types of MWE, an analysis as word-with-spaces might not be the most appropriate, as argued by many (\citet{Sagetal01} to give just one example, see Chapter \ref{relatedwork}).

\begin{table}[H]
	\footnotesize
	\def\arraystretch{1.20} 
	\centering
	\begin{tabular}{| l | l | l | c |} \hline
		\textbf{model}
		& \textbf{detector type}
		& \textbf{resolver type}
		& \textbf{F$_1$}
		\\ \hline
		A &  &  & 85.15\\ \hline
		B1 & exhaustive & longest & 85.18 \\
		B2 & exhaustive & leftmost & 85.02 \\
		B3 & Proper Nouns & longest & \textbf{85.28} \\
		B4 & Length 2 & leftmost & 85.07 \\
		B5 & Stop words & longest & 85.19 \\
		\hline
	\end{tabular}
	\caption{F$_1$-measure of unlabelled dependencies against gold standard A using different recognizers and the `from A' model combining method\label{tab:res3}}
\end{table}

\section{Conclusion}
\tab To the question of whether we can improve CCG parsing accuracy with information about MWEs obtained by an automatic recognizer we answered that yes, it seems that we can. We have shown that \modelB can outperform \modelA on many evaluation schemes. We further answered that there can be both a training and a parsing effect. We have however been unable to show that we can significantly beat \modelA on gold standard\subscript{A} and the improvements we have made are not dramatic. To the question of whether or not the kind of MWE recognition that is conducted makes a difference in improving the parsing model we answered that yes it does seem to make a difference. We have obtained very encouraging but limited results and we further discuss them in the next Chapter.

\chapter{Discussion}
\label{discussion}

\section{Introduction}
\tab In Chapter \ref{relatedwork}, we saw that Syntactic parsing is a problem that is far from being solved although it is a task well worth working on because it is central to NLP. We saw that working with lexicalized formalisms was an appropriate thing to do because they are linguistically motivated and working with them has shown encouraging results. We also saw that using MWE information for syntactic parsing was a useful direction. Two hypotheses were put forward: that CCG parsing can be improved by adding information about MWEs and that using different types of recognizers could show different pictures. In Chapter \ref{methodology} we refined the first of these hypothesis: we said we wanted to find out whether or not automatic MWE recognition can be useful to CCG parsing. We also decoupled this question into two subquestions: whether or not there is a parsing effect (the parser is helped in its decision by collapsed data) and whether or not there is a training effect (the parser learns something useful). We also presented an approach to testing these hypotheses which we conducted. We reported results in Chapter \ref{exp} which seemed to confirm both hypotheses although evidence to support each was not strong. In this chapter, I further discuss the results as well as discuss limitations in the methodology, the way in which they affect the results and the way in which they can be improved in future work.

\section{Data collapsing}
\begin{figure}[H]
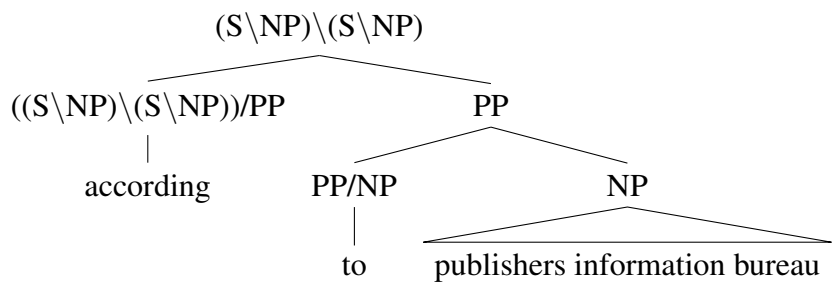

	\Tree [.(S\textbackslash NP)\textbackslash(S\textbackslash NP)
		[.((S\textbackslash NP)\textbackslash(S\textbackslash NP))/PP according ]
		[.PP
			[.PP/NP to ]
	\qroof{publishers information bureau}.NP ] ]
	\caption{Tree with MWE units that are not siblings\label{nstbis}}
\end{figure}

\begin{figure}[H]
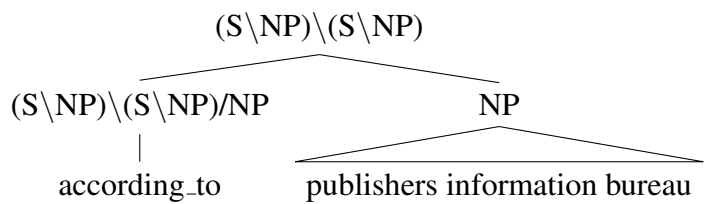

	\Tree [.(S\textbackslash NP)\textbackslash(S\textbackslash NP)
		[.(S\textbackslash NP)\textbackslash(S\textbackslash NP)/NP according\_to ]
	\qroof{publishers information bureau}.NP ]
	\caption{Ideal MWE non-sibling collapsing\label{nscbis}}
\end{figure}

\tab  Collapsing siblings showed a beneficial impact on parsing accuracy although by a margin that is not dramatic. This is not surprising since our collapsing algorithm does not modify much information in the tree. It only changes the probability distributions of the lexical categories of the nodes involved, i.e. children and parent. It does not modify the rest of the tree. One of the limitations that has been mentioned is the fact that we only collapsed MWEs when they are siblings in the data. An algorithm that would collapse non-siblings and return trees as described in Section \ref{treecoll} would make more radical changes to the tree and probably affect the probability model much more. In Chapter \ref{methodology}, an example of MWE for which units are not siblings in the Tree was given together with its ideal collapsed tree. Both Figures are reproduced in Figure \ref{nstbis} and \ref{nscbis} respectively for convenience. In this example, not only does the category of the leaf node `to' disappear (as is always the case in sibling collapsing) but its parent node `PP' also disappears (as never happens with sibling collapsing). In addition, a new category is created for according\_to whereas in the sibling case, the MWE just inherits the category of the parent node. More probability distributions would therefore get affected by this collapsing algorithm.

\tab It was mentioned that MWEs were more often siblings than not siblings but it was also mentioned that this actually depends on the MWE recognition method. It would therefore be interesting to see if there is an algorithm that can collapse all MWEs and if such an algorithm would in the end lead to a bigger impact on the parsing accuracy. It is however quite interesting to note that even in a limited setting as the one that we explored, there is a positive impact on parsing accuracy. Future work involves extending our current model to the non-sibling cases to see if the small but significant impact made by our incomplete method can be improved by a complete method.

\section{MWE recognition}
\tab A few experiments have been conducted on different MWE recognizers but because of time limitations we could not conduct a comprehensive study of the effect of each MWE recognition method on parsing accuracy. It seemed that using the resolver that chooses the longest matching MWEs worked better for our purposes than the one that chooses the leftmost MWE but more experiments would be required to study this and understand the different effects better. It seemed that using only Proper Nouns was the best MWE recognition type for our purposes but a more comprehensive study would include more experiments on different types of MWEs. Such a study would give better insight into the type of MWEs that are best collapsed as one lexical item and into the reasons why this is the case. In addition, although we used several layers of the MWE recognition library we were using by experimenting with the different tools, we only used a restricted set of data, the set of data offered by the library. There is a lot more MWE data available than this data set. A more comprehensive study on more data could lead to interesting results and bigger differences in the resulting parsing accuracy. In addition, MWEs have been said to be quite domain-dependent \citep{Sagetal01} and working on different data might impact the results differently.

\section{Evaluation}
\tab It has been said a few times that the fact that we did not have an algorithm to collapse non-sibling MWE units made the evaluation part difficult. We obtain the best results when we have gold information about siblings but these results cannot be fully attributed to our changed models. We implemented evaluation schemes to cross-verify the results obtained by testing our models on data where all MWEs are collapsed and we showed improvements of our models even on these evaluation schemes but with lower effects. However, when using these cross-validation schemes, we do not have reliable gold standards so our results might be downplaying bigger effects. In any case, showing improvements on all evaluation schemes shows that our models are useful but our imperfect evaluation might be hiding bigger effects.\\
\tab In addition, we used a purely quantitative evaluation but it would be very interesting to conduct a more qualitative evaluation such as the one by \citet{korkontzelos2010} described in Section \ref{korkontz}. This could help understand the cases in which the collapsed data lead to increased accuracy and the cases in which it leads to decreased accuracy.

\section{Overall approach}
\tab We answered both our questions positively but the improved parsing accuracy we obtained is not dramatic. The general approach could also be improved and possibly lead to better results. In Section \ref{choiceofapp} it was mentioned that we chose the approach of changing training and test data before parsing because it is a method that has proven useful in the past, because it is a sound baseline for future research and because of time limitations. It was mentioned that a more sophisticated approach is to not only change training and test data but also to change the parsing algorithm. This would be interesting future work. Another thing we could do which could also be interesting is to integrate MWE recognition in the collapsing algorithm. As a matter of fact, conflicts between MWEs are resolved without information about the parsing structure. When collapsing we can make the decision of discarding MWEs because MWE units are not siblings in the tree. These MWEs that are discarded however could embed smaller MWEs and it would be useful if we could access this information during collapsing. 

\section{Conclusion}
\tab We positively answered both our hypotheses that MWE information can help CCG parsing and that different MWE recognition models impact parsing accuracy differently but our approach could be improved in many ways: in the data collapsing, in the experiments on MWE recognition, in the evaluation and in the overall approach we took. We leave this to future research and will now try to see what can be concluded from the work undertaken here.

\chapter{Conclusion}
\label{ccl}
\paragraph*{Contributions}
Our main contributions in this work are:
\begin{itemize}
	\item Encouraging results on a notoriously hard task
	\item Improvements on CCG parsing with automatic MWE recognition
	\item Significant results despite limited settings
	\item An algorithm to automatically collapse MWEs in a treebank
	\item Techniques for distinguishing training from parsing effects
	\item Empirical support that there is both training and parsing effects
	\item Interesting differences in results when using different recognizers
\end{itemize}
\tab The task we have been trying to improve in this thesis is the task of syntactic parsing, i.e. assigning a structure to a sentence. The importance of working on this task for NLP applications was emphasized together with the difficulty of improving it. Arguments for working with a lexicalized grammar such as CCG were put forward, the most important of which being their linguistic plausibility and the fact that they are increasingly successfully used for the syntactic parsing task. Working on the lexicon by adding MWE information to it was singled out as a useful direction because it has proven useful in the past and because it seemed an interesting approach to attempt within the framework of a lexicalized grammar. We built on previous work which had shown the benefits of giving information about MWEs to a syntactic parser. It had been shown to work for deterministic dependency parsing, shallow parsing and deterministic constituency parsing but not for statistical constituency parsing. We implemented an existing pipeline which consists in collapsing MWEs in training and test data and adapted it to our purposes. We gave further evidence supporting these studies and showed that statistical constituency parsing with a lexicalized grammar too can benefit from MWE information. Our study provided further empirical support to the hypothesis that MWE information can improve syntactic parsing by showing that we can improve CCG parsing with information about MWEs. \\
\tab MWE identification was also identified as a notoriously difficult task although important for many applications because MWEs violate usual compositional rules and can be the source of many errors if not handled properly. We have shown that using an existing automatic recognizer as a source of MWE information was useful which had so far been left a bit unclear in the literature. \\
\tab Our results have shown small but significant improvements on previous models which is very encouraging given the difficult task at hand and given the restricted settings we have worked with. We have as a matter of fact hypothesized that the results were very much limited by the methodology used and have suggested ways of improving the current approach. Our biggest contributions however are not in the results we obtained but in the techniques we proposed. The study has proposed novel techniques to improve on previous pipelines. We have proposed an algorithm to automatically collapse MWEs in a treebank which can be used with other formalisms although this algorithm is limited to collapsing MWEs which form a constituent in the tree. More importantly, we have proposed ways of experimenting with our models in a way that we can distinguish parsing (the parser is helped in its decisions by collapsed data) from training effects (the parser learns something useful) in the evaluation and have shown evidence for both. In addition, we have proposed to turn the MWE recognition part of the pipeline into an experimental part and have shown that doing so leads to differences in results. This is especially interesting in that it is never quite clear in the literature what counts as an MWE. It is therefore interesting to discover what types of MWE our technique is best suited to and our method could be used to discover interesting properties of MWEs. \\
\paragraph*{Future work}
We propose the following for future work:
\begin{itemize}
	\item Extending the collapsing algorithm to the non-sibling case
	\item Testing more MWE recognition methods
	\item Using more and/or different data for MWE recognition
	\item Further integrating MWE recognition and syntactic parsing
	\item Integrating MWE recognition and our collapsing algorithm
	\item Conducting qualitative evaluation
\end{itemize}
\tab A lot more interesting research can still be done on the interaction between MWE identification and syntactic parsing. Theoretical research has emphasized the need to give different syntactic representations for different types of MWEs but a lot of empirical work is still needed if we want to automatically assign sensible syntactic representations to MWEs. In the meantime, we believe to have opened new perspectives in the study of the integration between syntactic parsing and MWE identification especially in relation to CCG parsing. We have given encouraging results on a difficult task and suggested ways of improving them. We have given further evidence that the integration of MWE identification with syntactic parsing is a promising and exciting research direction and we have shown that it is a research direction that still has a lot to offer.

\insertbibliography

\end{document}